\documentclass[conference]{IEEEtran}
\usepackage{cite}
\usepackage{amsmath,amssymb,amsfonts}
\usepackage{algorithmic}
\usepackage{graphicx}
\usepackage[round]{natbib}              
\usepackage{url}            
\usepackage{subcaption}
\usepackage{textcomp}
\usepackage{xcolor}
\usepackage{soul}
\usepackage{float}
\usepackage{geometry}
\usepackage{multirow}       
\usepackage{mathtools}      
\usepackage{fancyhdr}       
\usepackage{orcidlink}      
\usepackage{wrapfig}        
\usepackage{hyperref}       
\hypersetup{
  colorlinks= true, 
  urlcolor= magenta, 
  linkcolor= red, 
  citecolor= blue 
}
\usepackage{lipsum}         

\def\BibTeX{{\rm B\kern-.05em{\sc i\kern-.025em b}\kern-.08em
    T\kern-.1667em\lower.7ex\hbox{E}\kern-.125emX}}

\makeatletter

\def\modelname{SambaMixer}

\title{\modelname: State of Health Prediction of Li-ion Batteries using Mamba State Space Models}

\author{José Ignacio Olalde-Verano \orcidlink{0000-0001-8058-156X}$^{*}$ , Sascha Kirch \orcidlink{0000-0002-5578-7555}$^{*, \dagger}$, Clara Pérez-Molina \orcidlink{0000-0001-8260-4155}$^{*}$, Sergio Martin \orcidlink{0000-0002-4118-0234}$^{*}$ \\
\normalsize \small{ $^{*}$UNED - Universidad Nacional de Educación a Distancia, Madrid, Spain}\\
\normalsize \small{ $^{\dagger}$Corresponding Author}\\
\normalsize \small{\{\href{mailto:jolalde5@alumno.uned.es}{jolalde5}, \href{mailto:skirch1@alumno.uned.es}{skirch1}\}@alumno.uned.es,
\{\href{mailto:clarapm@ieec.uned.es}{clarapm}, \href{mailto:smartin@ieec.uned.es}{smartin}\}@ieec.uned.es} \\
}

\begin{document}

\maketitle

\begin{abstract}
The state of health (SOH) of a Li-ion battery is a critical parameter that determines the remaining capacity and the remaining lifetime of the battery.

In this paper, we propose \modelname{} a novel structured state space model (SSM) for predicting the state of health of Li-ion batteries. The proposed SSM is based on the MambaMixer architecture, which is designed to handle multi-variate time signals. 

We evaluate our model on the NASA battery discharge dataset and show that our model outperforms the state-of-the-art on this dataset. 

We further introduce a novel anchor-based resampling method which ensures time signals are of the expected length while also serving as augmentation technique. Finally, we condition prediction on the sample time and the cycle time difference using positional encodings to improve the performance of our model and to learn recuperation effects. Our results proof that our model is able to predict the SOH of Li-ion batteries with high accuracy and robustness.

\end{abstract}
\begin{IEEEkeywords}
Li-ion battery, mamba, state space model, state of health prediction, multi-variate time series, deep learning
\end{IEEEkeywords}

\section{Introduction}\label{sec:introduction}
Lithium-ion (Li-ion) batteries are among the most widely used energy storage solutions today, powering everything from consumer electronics to electric vehicles (EVs), even resulting in the 2019 Nobel Price in Chemistry \citep{fernholm_nobel_2019}. Their popularity stems from their high energy density, long lifespan, and low self-discharge rate, which make them both efficient and durable \citep{li_30_2018}. 

However, ensuring safety, reliability, and efficiency of Li-ion batteries over time requires sophisticated battery management systems (BMS) that monitor, control, and optimize battery performance. Accurate prediction of either the state of health (SOH) or state of charge (SOC) are essential to prevent unexpected failures and extend battery life. 

Traditional BMS often rely on equivalent circuit models (ECM) \citep{liu_comparative_2014} as well as electrochemical models (EM) \citep{elmahallawy_comprehensive_2022}, but these are limited by their complexity and sensitivity to varying operational conditions. In recent years, deep learning models have emerged as powerful tools for health prediction in Li-ion batteries due to their ability to learn complex, non-linear relationships directly from data, providing more accurate, adaptive, and scalable solutions for real-time health monitoring.

We noticed that most of recent works are not considering recent advances of deep learning \citep{mazzi_lithium-ion_2024, Yao2024}. We acknowledge that some works \citep{Crocioni2020} have put their focus on deploying models on embedded devices to show that small deep learning based models can be used for real-time health monitoring of Li-ion batteries. At the same time the problem of SOH prediction is a multi-disciplinary problem that requires expertise in many different disciplines like battery technology, signal processing, and deep learning. Some works use modern transformer architectures \citep{Feng2024,Gomez2024,Zhu2024}, which have shown great success in many deep learning disciplines like natural language processing and computer vision. While these show great performance, they are not well-suited for time series data with many measurement samples due to their quadratic work complexity \citep{keles2022computationalcomplexityselfattention} and require a substantial amount of resources to train and large datasets to converge \citep{Popel_2018}.\\

In this paper we propose \modelname{}, a novel deep learning model based on Mamba state space models \citep{gu_mamba_2024,behrouz_mambamixer_2024} for predicting the SOH of Li-ion batteries. Our model is designed to handle long-range temporal dependencies in time series data and passing information between channels in multi-variate time series data. We evaluate our model NASA's real-world dataset of Li-ion battery discharge cycles \citep{saha2007nasa} and demonstrate its superior performance compared to state-of-the-art deep learning models. \\

In this sense, we summarize our main contributions of this paper as follows:
\begin{enumerate}
  \item Introducing Mamba state space models to the problem of Li-ion battery SOH prediction.
  \item Developing an anchor-based resampling scheme to resample time signals to have the same number of samples while serving as a data augmentation method.
  \item Applying a sample time-based positional encoding scheme to the input sequence to tackle sample jitter, time signals of varying length and recuperation effects of Li-ion batteries.
\end{enumerate}

We release our code on GitHub\footnote{GitHub Repo: \url{https://github.com/sascha-kirch/samba-mixer}}.
\section{Related Work} \label{sec:related_work}
\subsection{State-of-Health Prediction of Li-ion Batteries} \label{subsec:related_work_soh}
\cite{Ren2023} categorizes battery SOH prediction methods into two classes: model-driven and data-driven methods. In this work we focus on data-driven methods.

Many works combine recurrent networks and convolution networks to predict a battery's SOH. \cite{mazzi_lithium-ion_2024} use a 1D-CNN followed by BiGRU layers, utilizing measured voltage, current, and temperature signals from the NASA PCoE dataset \citep{saha2007nasa}. Utilizing the same dataset, \cite{Yao2024} develop a CNN-WNN-WLSTM network with wavelet activation functions. \cite{Shen2023} use an extreme learning machine (ELM) algorithm on voltage signals measured during charging mode. \cite{Wu2022} combine convolutional and recurrent autoencoders with GRU networks. \cite{Zhu2022} use a CNN-BiLSTM with attention for SOH and remaining useful life (RUL) estimation. \cite{Ren2021} employ an autoencoder feeding parallel CNN and LSTM blocks. \cite{Tong2021} develop an ADLSTM network with Bayesian optimization. \cite{Tan2020} propose a feature score rule for LSTM-FC networks. \cite{Crocioni2020} compare CNN-LSTM and CNN-GRU networks. \cite{Li2020} introduce an AST-LSTM network. \cite{Yang2020} merge CNN with random forest in a CNN-RF network. \cite{Garse2024} use a random forest regression and FC network in the RFR-ANN model. \cite{Chen2024} tackle SOH with a self-attention knowledge domain adaptation network.

Other works focus on transformer-based models. \cite{Feng2024} introduce GPT4Battery, a large language model (LLM) finetuned to estimate SOH on the GOTION dataset \citep{lu_deep_2023}. It employs a pre-trained GPT-2 backbone, followed by a feature extractor and two heads for charging curve reconstruction and SOH estimation. \cite{Gomez2024} use a temporal fusion transformer (TFT) on a Toyota dataset \citep{severson_data-driven_2019}, integrating Bi-LSTM layers for time series forecasting. \cite{Zhu2024} develop a Transformer with sparse attention and dilated convolution layers on the CALCE \citep{he_prognostics_2011} and NASA PCoE datasets. \cite{Huang2024} use singular value decomposition before inputting data into a Transformer model. \cite{Nakano2024} combine a CNN with a Transformer model in an experimental EV. They feed voltage, current, and speed signals along with the SOC.

\subsection{Structured State Space Models} \label{subsec:related_work_ssm}

Recently, state space models (SSMs) made their debut in the field of deep learning challenging the dominance of transformers \citep{vaswani_attention_2017} in sequential data tasks. While the transformer is successfully used in most fields of deep learning, its quadratic scaling law makes it challenging and expensive to be used for certain tasks with long sequences.

\cite{gu_lssl_2021}'s LSSL model incorporated \cite{gu_hippo_2020}'s HiPPO Framework into SSMs and showed that SSMs can be trained. They further highlighted the duality of its recurrent and convolution representation, which meant, that it can be inferred with $O(N)$ complexity in its recurrent view and trained in parallel leveraging modern hardware accelerators using the convolution representation. The S4 model by \cite{gu_s4_2022} further employed a certain structure upon its state matrix A, which allowed for a more efficient construction of the convolution kernel required for training. Many subsequent work \citep{smith_s5_2023,gupta_dss_2022,gu_s4d_2022,fu_h3_2023,gu_how_2022} further improved upon existing SSMs which ultimately led to the development of the Mamba model by \cite{gu_mamba_2024}. Mamba added selectivity into the SSM increasing its performance while still featuring sub-quadratic complexity during inference. It is this Transformer-like performance while scaling sub-quadratically with the sequence length which makes it especially suited for sequential data tasks with long sequences such as audio \citep{lin_audio_2024,erol_audio_2024}, images \citep{nguyen_s4nd_2022,liu_vmamba_2024,zhu_vision_2024}, video \citep{chen_video_2024,li_videomamba_2024}, NLP \citep{lieber_jamba_2024}, segmentation \citep{wan_sigma_2024}, motion generation \citep{zhang_motion_2024} and stock prediction \citep{shi_mambastock_2024}.
Recent work focuses on the connection between attention and SSMs \cite{ali_hidden_2024,dao_mamba-2_2024} to simplify its formulation and to be able to leverage the vast amount of research done on attention mechanisms of transformers and its hardware aware and efficient implementations. \cite{behrouz_mambamixer_2024} extends Mamba-like models to apply its selectivity not only along tokens but also along channels, making it especially well suited for multi-variate time signals such those found in the state of health prediction of Li-ion batteries.

\section{Preliminaries}\label{sec:preliminaries}

\subsection{State-of-Health of Li-ion Batteries}\label{subsec:preliminaries_soh}
Lithium-ion (Li-ion) batteries are widely used in portable electronics, electric vehicles, and renewable energy storage systems due to their high energy density, long cycle life, and low self-discharge rate. The degradation of the battery's performance is often shown by the battery's state of health (SOH) which decreases over time as a result of a variety of internal and external factors which we will detail later in this section. 
The SOH of a battery is a measure of its ability to deliver the rated capacity and power compared to its initial state. 

The state of health $SOH_{k}\,[\%]$ of a Li-ion battery in percentage is defined as
\begin{equation}
    \label{eq:soh}
    SOH_{k}[\%] = \frac{Q_{k}}{Q_{r}} \cdot 100,
\end{equation}
where $Q_{k}$ is the battery's current capacity at cycle $k$ and $Q_{r}$ its rated capacity.

\begin{figure*}[t]
    \centering
    \includegraphics[width=\textwidth]{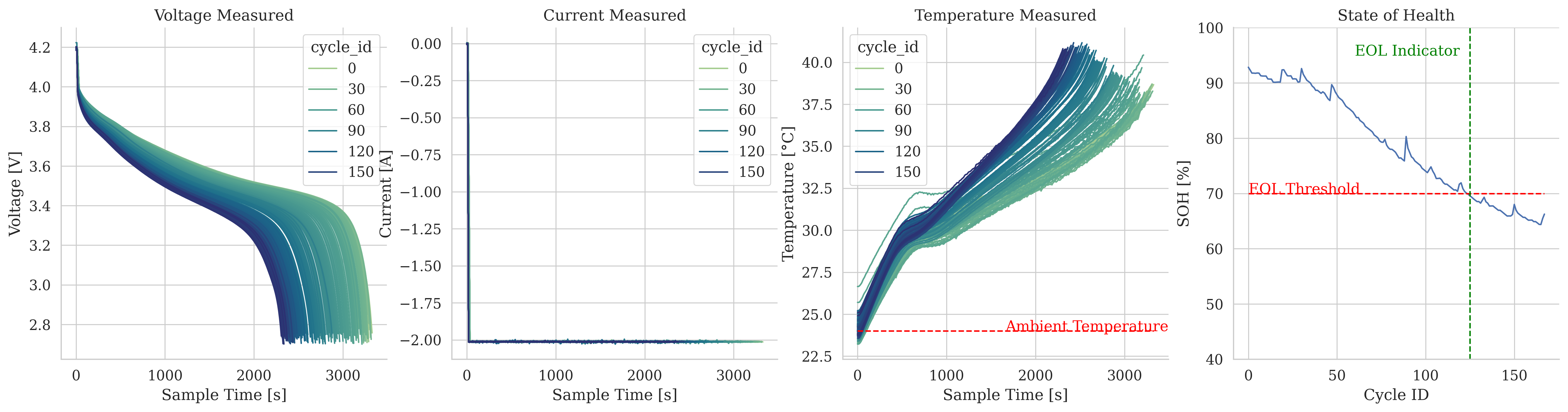}
    \caption{Effect of battery aging on the measured voltage, current and temperature of various discharge cycles of a Li-ion battery. Battery \#5 of NASA's battery dataset \citep{saha2007nasa}.}
    \label{fig:battery_aging}
\end{figure*}

As the battery is used and repeatedly charged and discharged, its SOH decreases with each cycle, which can be observed in the measured voltage, current and temperature profiles. Figure \ref{fig:battery_aging} depicts an example.

The EOL of a battery is defined as the point at which the battery can no longer deliver the rated capacity and power and is considered to be at the end of its useful life. The EOL of a battery is typically reached when the SOH of the battery drops below a certain threshold, e.g., 70\% of the rated capacity. It is important to note that due to recuperation effects, the SOH of a battery can increase again hence passing the EOL threshold multiple times. In this work, we set the EOL indicator to the first cycle after the SOH drops below the threshold for the last time.

As previously stated, there are internal factors and external factors that contribute to the aging of Li-ion batteries \citep{Liu2023}. Internal factors are concerned with the chemical properties and external factors with for example manufacturing, environment and the usage of the battery, to name a few.

\subsubsection{Internal Factors}\label{subsec:preliminaries_internal_factors}
\cite{Zeng2023} identifies 21 possible internal factors causing a degradation in a Li-ion battery's state of health. 
These factors can be grouped into three fundamental concepts: loss of lithium inventory (LLI), loss of active material (LAM) and increase in internal resistance. Within these three groups, the loss of lithium inventory is one of the most impactful on the aging process \citep{Li2019}.

LLI factors include lithium precipitation and SEI formation. Lithium precipitation occurs at the anode during charging, where lithium ions form dendrites that can puncture the separator, causing short circuits \citep{Yang2017}. SEI formation happens during the first charge, reducing available lithium ions and affecting their dynamics \citep{Kekes2016}.

LAM factors primarily involve lithium oxide degradation at the cathode, leading to gas generation and increased internal resistance \citep{Wang2021}.

Increased internal resistance is also caused by electrode corrosion \citep{Yamada2020}, electrolyte decomposition \citep{Wang2012}, and diaphragm degradation \citep{Yang2016}.

\subsubsection{External Factors}\label{subsec:preliminaries_external_factors}
External factors are categorized based on the battery's temperature, charge rate, overcharge/overdischarge level and mechanical stresses \citep{Tian2020, Vetter2005}. 

Using a battery outside its specified temperature range, too high and too low temperatures can both affect the battery's performance in different ways. High temperatures can lead to the formation of solid electrolyte interface, degradation of the cathode, and ultimately thermal runaway \citep{Waldmann2014, Finegan2015}. Too low temperatures slow down the transport of lithium ions, increase internal resistance, and affect the battery's capacity \citep{Zichen2021}.

Charging a battery at a high rate, meaning with high charging current, can lead to the precipitation of ions on the anode, which is favored by the increase in temperature due to the Joule effect \citep{Gao2017, Jaguemont2016}. Similarly, overcharging a battery can lead to irreversible structural changes in the cathode and an increase in internal resistance \citep{He2011, Ouyang2015}. Overdischarging a battery can result in the dissolution of the anode material into $Cu$ ions, which can generate dendrites in the charging process \citep{Yamada2020}.

To conclude, a vast number of internal and external factors can contribute to the degradation of a Li-ion battery's state of health, making it a complex and challenging problem to model.

\subsection{Structured State Space Models}\label{subsec:preliminaries_ssm}
A state space model (SSM) describes the relationship between an input signal $x(t)$ and an output signal $y(t)$ through a hidden state $h(t)$, which evolves over time according to a linear dynamical system. The SSM is defined by the following equations:
\begin{align}\nonumber
    h'(t) &= \mathbf{A} h(t) + \mathbf{B} x(t), \\ \label{eq:ssm_continuous_time}
    y(t) &= \mathbf{C} h(t) + \mathbf{D} x(t).
\end{align}

Matrix $\mathbf{D}$ transforms the input $x(t)$ directly to the output $y(t)$ and is usually pulled from the SSM and modeled as a skip connection.
Since most applications deal with discrete signals (e.g. discretized analog time signals or text tokens) and the fact that the above differential equation is not directly solvable, the SSM is discretized, resulting in the following discrete-time SSM:
\begin{align}\nonumber
    h_t &= \bar{\mathbf{A}} h_{t-1} + \bar{\mathbf{B}} x_t,\\ \label{eq:ssm_discrete_time}
    y_t &= \mathbf{C} h_t,
\end{align}

where $\bar{\mathbf{A}}$ and $\bar{\mathbf{B}}$ are the discretized state matrix and input matrix, respectively. Many discretization techniques have been applied, with the ZOH (Zero order hold) discretization technique being the most prominent one in recent works:
\begin{align}\nonumber
    \bar{\mathbf{A}} &= e^{\boldsymbol{\Delta} \mathbf{A}},\\ \label{eq:zoh}
    \bar{\mathbf{B}} &= \left( \boldsymbol{\Delta} \mathbf{A} \right)^{-1} \left(\bar{\mathbf{A}}- I \right) \: \boldsymbol{\Delta} \mathbf{B}.
\end{align}

In other words, the discrete SSM maps an input sequence $x \in \mathbb{R}^{L \times D} = \{x_t| t \in \mathbb{N}_L\}$ to an output sequence $y \in \mathbb{R}^{L \times D} = \{y_t| t \in \mathbb{N}_L\}$ with $\mathbb{N}_L$ being the indices of the sequence with $L$ samples and $D$ the dimensionality of individual data points. Since matrices $\bar{\mathbf{A}}$, $\bar{\mathbf{B}}$ and $\mathbf{C}$ are constant over time, the SSM is said to be a linear time-invariant (LTI) system. In an LTI system, the recurrent representation of the SSM can be written in form of a convolution:
\begin{align}\nonumber
    \bar{\mathbf{K}} &= \left( {\mathbf{C}} \bar{\mathbf{B}}, {\mathbf{C}} \bar{\mathbf{A}} \bar{\mathbf{B}}, \dots, {\mathbf{C}} \bar{\mathbf{A}}^{L-1} \bar{\mathbf{B}} \right),\\ \label{eq:ssm_convolution_representation}
    y &= x \ast \bar{\mathbf{K}}.
\end{align}

Note that the convolution kernel $\bar{\mathbf{K}}$ is a function of the SSM matrices and contains $L$ elements, which is quite expensive to compute for large $L$ and dense matrices $\bar{\mathbf{A}} \in \mathbb{R}^{N \times N}$. \cite{gu_s4_2022} restricted matrix $\mathbf{A}$ to be a diagonal plus low rank (DPLR) matrix with $\mathbf{A}=\Lambda-PP^{*}$, which allows for a more efficient computation of of the convolution kernel $\bar{\mathbf{K}}$.

To further increase the performance of the SSM, \cite{gu_mamba_2024} presented Mamba which added selectivity to the SSM, by making matrices $\mathbf{B}_{t}$, $\mathbf{C}_{t}$ and $\mathbf{\Delta}_{t}$ time-variant, meaning each token is processed by its own matrix.

\cite{behrouz_mambamixer_2024} highlighted that Mamba's selectivity only applies on token level, but not on channel level, meaning information cannot be passed between channels. To address this issue, they proposed the MambaMixer, which adds channel-wise selectivity to the SSM, making it well suited for multi-channel data such as images or multi-variate time series.

A little simplified, the MambaMixer consists of two mixing operations, the token mixer $M_{\texttt{token}}$ and the channel mixer $M_{\texttt{channel}}$, which are defined as follows:
\begin{align}\nonumber
    M_{\texttt{token}} & : \mathbb{R}^{L \times D} \mapsto \mathbb{R}^{L \times D}, \\
    M_{\texttt{channel}} & : \mathbb{R}^{D \times L} \mapsto \mathbb{R}^{D \times L}.
\end{align}

Those mixers are build from one or more Mamba-like blocks. To obtain the output $y$ of a single MambaMixer block, the input $x$ is first processed by the token mixer $M_{\texttt{token}}$ and then by the channel mixer $M_{\texttt{channel}}$:
\begin{align}\nonumber
    y_{\texttt{token}}   &= M_{\texttt{token}}(x_{\texttt{token}}), \\ \nonumber
    y_{\texttt{channel}} &= M_{\texttt{channel}}(x_{\texttt{channel}}^T), \\
    y                    &= y_{\texttt{channel}}^T.
\end{align}

Note that the transpose operation is necessary to make the channel mixer work on the channel dimension.

Inspired by DenseNet \citep{huang2018denselyconnectedconvolutionalnetworks}, MambaMixer further implements a learned weighted averaging of earlier blocks' outputs to the current block's input, which is defined as follows:
\begin{align} \nonumber \label{eq:preliminaries_weighted_average}
    {x}_{\texttt{token}}^{(m)} &=  \sum_{i = 0}^{m - 1} \alpha_{m}^{(i)} \: y^{(i)}_{\texttt{token}} &+ \sum_{i = 0}^{m - 1} \beta_{m}^{(i)} \: y^{(i)}_{\texttt{channel}}, \\
    {x}_{\texttt{channel}}^{(m)} &=  \sum_{i = 0}^{m} \theta_{m}^{(i)} \: y^{(i)}_{\texttt{token}} &+ \sum_{i = 0}^{m - 1} \gamma_{m}^{(i)} \: y^{(i)}_{\texttt{channel}},
\end{align}
where $m$ is the current index of the $M$ stacked MambaMixer blocks,  $\alpha_{m}^{(i)}$, $\beta_{m}^{(i)}$, $\theta_{m}^{(i)}$, and $\gamma_{m}^{(i)}$ are learnable parameters and $y^{(0)}_{\texttt{token}} = y^{(0)}_{\texttt{channel}} = x_{\texttt{embedd}}$, where $x_{\texttt{embedd}}$ is the input to the encoder model.

\section{Proposed Method}\label{sec:proposed_method}

\subsection{Problem Formulation}\label{subsec:proposed_method_problem_formulation}
Let $\mathbb{N}_B=\{0, 1, \dots, \Psi-1\}$ be the indices of $\Psi$ different Li-ion batteries $B=\{b_{\psi} | \psi \in \mathbb{N}_B\}$ and $\mathbb{N}_{K}^{\psi}=\{0, 1, \dots, K^{\psi}-1\}$ be the indices of $K^{\psi}$ different discharge cycles $C^{\psi}=\{k| k \in \mathbb{N}_{K}^{\psi}\}$ for each of the $\Psi$ different Li-ion batteries in $B$. Each discharge cycle $k$ consists of a sequence of measured samples of the current signal $I_{k}$, voltage signal $V_{k}$, temperature signal $T_{k}$ and sample time $S_{k}$. All signals are measured at the battery's terminal.
\begin{equation}
    I_{k} = \{i_{t}^{(k)}\},V_{k} = \{v_{t}^{(k)}\}, T_{k} = \{\tau_{t}^{(k)}\}, S_{k} = \{s_{t}^{(k)}\},
\end{equation}
where $t \in [0, L_{k}^{\psi}) \subset \mathbb{N}$ is the index of individual samples, with $L_{k}^{\psi}$ being the total number of samples in cycle $k$ of battery $b_{\psi}$. Note that $S_{k}$ is the sample time in seconds, where $s_{t=0}^{(k)}$ always starts at 0\,s.

Through our anchor-based resampling introduced in section \ref{subsubsec:proposed_method_anchor_based_resampling} we ensure that for all cycles in $C^{\psi}$ the total number of samples are equal $L_{k}^{\psi} = L$.

By concatenating the input signals, we get the input tensor $P_{k} \in \mathbb{R}^{L \times 4}$ for cycle $k$ of battery $b_{\psi}$:
\begin{equation}
    P_{k} = I_{k} \mathbin\Vert V_{k} \mathbin\Vert T_{k} \mathbin\Vert S_{k},
\end{equation}

where $\mathbin\Vert$ denotes the concatenation operation. The objective of \modelname{} is to learn a parameterized function $f_{\Theta}$ that maps the input tensor $P_{k}$ to the state of health $SOH_{k}$ for a given cycle $k$ of a given battery $b_{\psi}$:
\begin{equation}
    f_{\Theta}: P_{k} \mapsto SOH_{k}.
\end{equation}

\subsection{The \modelname{} Model Architecture}\label{subsec:proposed_method_architecture}

\begin{figure*}[t]
\centering
\includegraphics[width=0.8\textwidth]{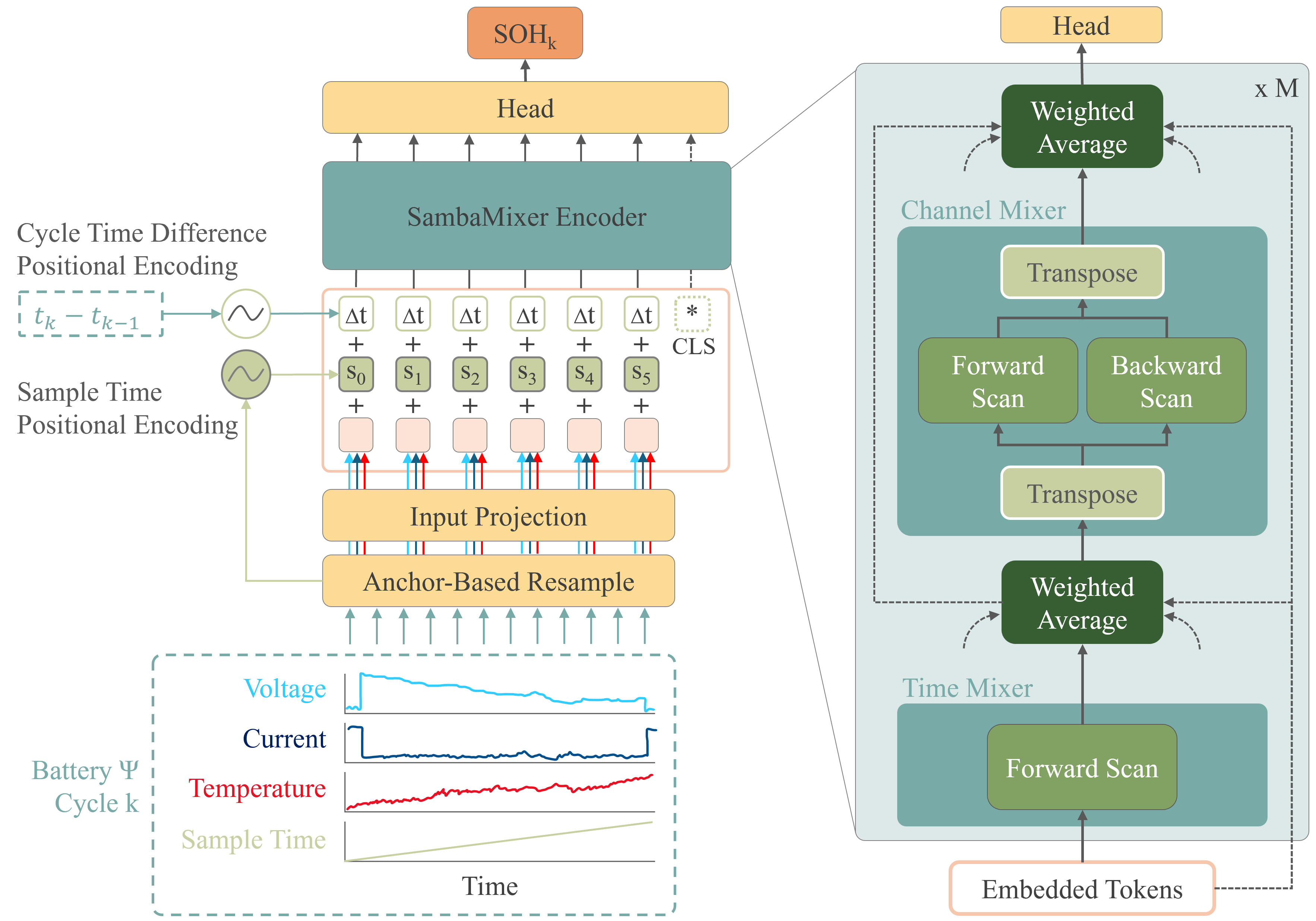}
\caption{\modelname{} architecture. We input a multi-variate time series of current, voltage, temperature and sample time. We first first resample the time signals using our anchor-based resampling technique. We then feed the resampled sample time into the sample time positional encoding layer. We further feed the time difference between two discharge cycles in hours into the cycle time difference positional encoding layer. The other signals, i.e. current, voltage and temperature are fed into the input projection. The projected signals are added to the sample time embeddings and the cycle time difference embeddings. Optionally, a CLS token can be inserted at any position. The embedded tokens are then fed into the \modelname{} Encoder. The \modelname{} Encoder consists of $M$ stacked \modelname{} Encoder blocks. The output of the encoder is finally fed into the head, which predicts the state of health of the current cycle $k$ for battery $b_{\psi}$.}
\label{fig:architecture}
\end{figure*}

A top-level view of our \modelname{}'s model architecture is depicted in Fig. \ref{fig:architecture}. It consists of five main components: Resampling, input projection, position encoding, encoder backbone and the prediction head.

We input a multi-variate time series of current, voltage, temperature and sample time of a single discharge cycle $k$ of a single battery $b_{\psi}$. Our \modelname{} model then predicts the state of health $SOH_{k}$ for that cycle.

\subsubsection{Anchor-Based Resampling of Time Signals}\label{subsubsec:proposed_method_anchor_based_resampling}

As said earlier, we use the discharge cycles of a battery to determine its state of health. Since those cycles become shorter with the battery aging and because different sample rates are chosen to sample the data, the number of samples from different discharge cycles and batteries vary drastically. Further, more samples result in a wider model which consequently also means more resources are required to train it. Depending on the discharge mode, the required number of samples varies a lot. For example, in a constant current discharge mode, the current is nearly constant and the voltage drops continuously. Hence, a few number of samples might suffice. On the other hand, high frequency discharge profiles might require more samples to avoid anti-aliasing effects and to be able the model the dynamics of the systems.

To conclude, there are many reasons why we need to be able to change the number of samples. We resample and interpolate the time signals to ensure we always have the same number of samples, using our anchor-based resampling technique.

Generally speaking, we define a resampling function $f_R$ that resamples the sample time sequence $S_k$ of length $L_{k}^{\psi}$. $L_{k}^{\psi}$ varies for each cycle $k$ and battery $b_{\psi}$. The result is the resampled sample-time sequence $S_{k}^{*}$ which has the the same length $L$ for all cycles and batteries.
\begin{equation}
    f_R: S_{k} \in \mathbb{R}^{L_{k}^{\psi}} \mapsto S_{k}^{*} \in \mathbb{R}^{L}.
\end{equation}

Once we have $S_{k}^{*}$, we linearly interpolate the current, voltage and temperature signal.

We experiment with three different approaches for the resampling function $f_R$: linear resampling, random resampling and our anchor-based resampling. Results are presented in section \ref{subsubsec:ablation_resampling}.

For the linear resampling $f_R^{l}$, we simply take $L$ equidistant samples between the min and max value of $S_k$.
\begin{equation}
    f_R^{l}(S_k) := linspace(\min(S_k), \max(S_k), L).
\end{equation}

For the random resampling $f_R^{r}$, we draw $L$ samples from a uniform distribution $\mathcal{U}$.
\begin{equation}
    f_R^{r}(S_k) := \{ s_t^{k}\}_{t=0}^{L}, \, with \, s_t^k \sim \mathcal{U}_{[\min(S_k), \max(S_k)]}.
\end{equation}

\begin{figure}[t]
    \includegraphics[width=\linewidth]{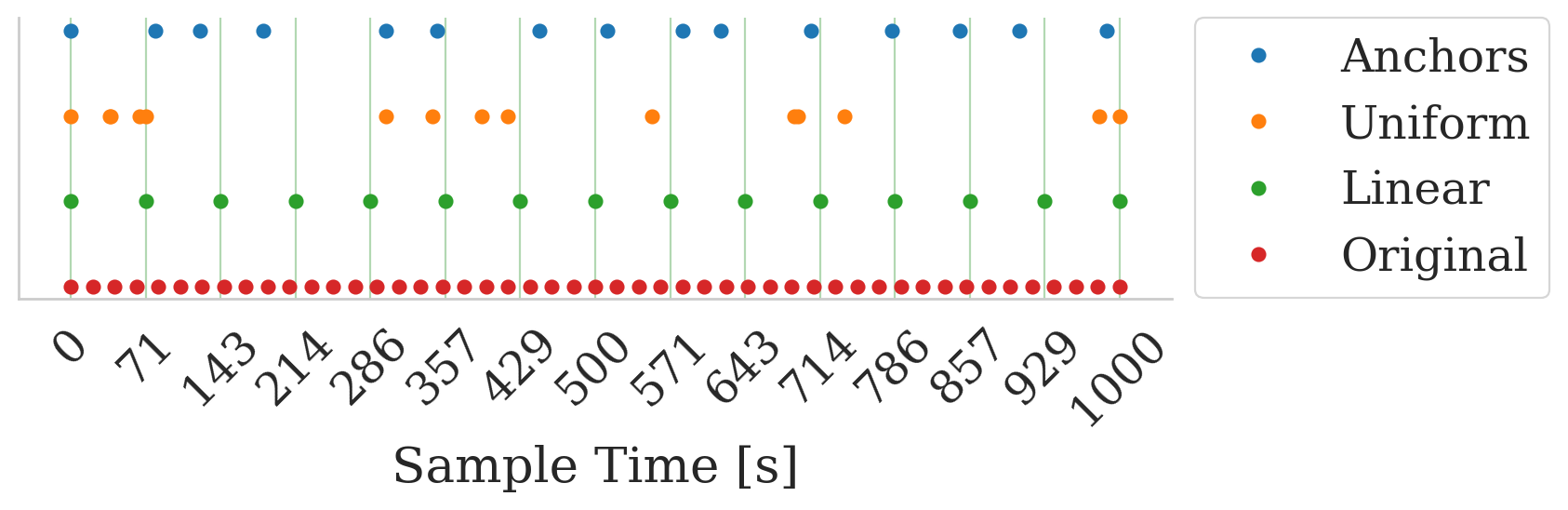}
    \caption{Resample techniques. Original: The original sample time sequence with $L_{k}^{\psi}$ samples. Linear: linear resampling with $L$ equidistant samples. Random: random resampling with $L$ samples drawn from a uniform distribution. Anchor: anchor-based resampling with random uniform noise $z$ added to $L$ equidistant samples.}
    \label{fig:resample}
\end{figure}

For our proposed anchor-based resampling $f_R^{a}$, we first define the anchors by using linear resampling $f_R^{l}$ and then add some noise $z$ to each anchor.
\begin{equation}
    f_R^{a}(S_k) := f_R^{l}(S_k) + \{z_t\}_{t=0}^{L}, \, with \, z_t \sim \mathcal{U}_{[-\frac{w}{2}, \frac{w}{2}]},
\end{equation}

where $w$ is the interval width between two linearly resampled samples.
In Figure \ref{fig:resample} we illustrate the resulting sample time for those three resample techniques.

\subsubsection{Input Projection}\label{subsubsec:proposed_method_input_projection}
We feed the resampled voltage, current, and temperature signals into our input projection. We use a simple linear projection layer to project the multi-variate time signal of $\mathbb{R}^{L \times 3}$ into $\mathbb{R}^{L \times d_{\texttt{model}}}$.

\subsubsection{Sample Time Position Embeddings}\label{subsubsec:proposed_method_sample_time_position_embeddings}
As shown in our top-level architecture in Fig. \ref{fig:architecture}, we use time information in our positional encoding layer to obtain position embeddings $PE^{(k)} \in \mathbb{R}^{L \times d_{\texttt{model}}}$ for cycle $k$ that are then added to the projected tokens.

In the original transformer by \cite{vaswani_attention_2017}, position embeddings were added since the transformer would otherwise has no knowledge of the order if its inputs because it has neither recurrence nor any convolutions. Among many possible techniques to either encode absolute or relative position, the sinusoidal position embedding like introduced by the transformer is still frequently used. It encodes the samples depending on their absolute position $p$ in the sequence.
\begin{align} \nonumber\label{eq:proposed_method_sinusodial_pe}
    & PE_{orig}\,[p, 2i]  & = \sin\left(p / 10.000^{2i/d_{\texttt{model}}}\right), \\
    & PE_{orig}\,[p, 2i+1] & = \cos\left(p / 10.000^{2i/d_{\texttt{model}}}\right).
\end{align}

An SSM on the other hand is a recurrent model and inside the Mamba block we also have a convolution. Even so, in VisionMamba by \cite{zhu_vision_2024}, position embeddings were still added to make sense of the spatial position of image patches. In this work, even though having a SSM applied on causal time signals, we still add position embeddings. 

Instead of encoding the position of the sample like in equation \ref{eq:proposed_method_sinusodial_pe}, we encode the sample time $s_{t}^{(k)}$ of cycle $k$ at position $p$ resulting in the positional embeddings $PE_{st}^{(k)}$.
\begin{align} \nonumber\label{eq:proposed_method_sample_time_pe}
    & PE_{st}^{(k)}\,[p, 2i]   & = \sin\left(s_{t=p}^{(k)} / 10.000^{2i/d_{\texttt{model}}}\right), \\
    & PE_{st}^{(k)}\,[p, 2i + 1] & = \cos\left(s_{t=p}^{(k)} / 10.000^{2i/d_{\texttt{model}}}\right).
\end{align}

Because we resampled the time signals to be all of equal length $L$, the distance between two samples is constant even though the sample time for the same position in different cycles $k$ of different batteries $b_{\psi}$ might be different.

The choice of our sample time based position encoding can be interpreted as an additional condition to the model, allowing it to learn from temporal information (e.g. how long it takes to discharge a battery) and making it robust against different sample rates and number of samples. 

Further, Li-ion batteries recuperate their capacity over time if not used. This means that the SOH of a cycle $k$ is not only dependent on the start time $t^{(k)}$ of the current cycle $k$, but also on the time difference $\Delta t^{(k)}$ in hours to the start time $t^{(k-1)}$ of the previous cycle $(k-1)$. 
\begin{equation}
    \Delta t^{(k)}:= t^{(k)} - t^{(k-1)}.
\end{equation}

We therefore add a second positional encoding to encode the time difference $\Delta t^{(k)}$ in hours between the start time $t^{(k)}$ of the current discharge cycle $k$ and the start time $t^{(k-1)}$ of the previous cycle $(k-1)$ so that the model can learn the recuperation of the battery's capacity over time. We obtain the positional embeddings $PE_{\Delta}^{(k)}$ for cycle $k$ at position $p$ as follows:
\begin{align} \nonumber\label{eq:proposed_method_sample_time_pe}
    & PE_{\Delta}^{(k)}\,[p, 2i]   & = \sin\left(\Delta t^{(k)} / 10.000^{2i/d_{\texttt{model}}}\right), \\
    & PE_{\Delta}^{(k)}\,[p,2i+1] & = \cos\left(\Delta t^{(k)} / 10.000^{2i/d_{\texttt{model}}}\right).
\end{align}

Our final positional embedding $PE^{(k)}$ for cycle $k$ is then the sum of the sample time positional embedding $PE_{st}^{(k)}$ and the cycle time difference positional embedding $PE_{\Delta}^{(k)}$:
\begin{equation}
    PE^{(k)} = PE_{st}^{(k)} + PE_{\Delta}^{(k)}.
\end{equation}

Note that the cycle time difference positional embedding $PE_{\Delta}^{(k)}$ is constant within a single cycle $k$ while the sample time positional embedding $PE_{st}^{(k)}$ is different for each sample $t$ in the cycle $k$.

We ablate different positional encoding methods in section \ref{subsubsec:ablation_positional_encoding}.

\subsubsection{Encoder Backbone}\label{subsubsec:proposed_method_encoder_backbone}
Our \modelname{} encoder backbone is strongly inspired by the TSM2 network of \cite{behrouz_mambamixer_2024}, which is a MambaMixer applied on time-series data. Since \cite{behrouz_mambamixer_2024} did not yet publish their implementation, we did implement it from scratch and give it the name \modelname{}.

We stack $M$ \modelname{} blocks to obtain our \modelname{} encoder. The \modelname{} consists of a Time Mixer module and a Channel Mixer module, which both consists of one or more Mamba SSM layers with different scan directions. The Time Mixer module applies the SSM along the token axis. It consists of a single forward scanning SSM due to the causal nature of sequence data. The Channel Mixer module on the other hand, does apply its SSMs on the channel/feature axis, which does not has this causal nature, hence we apply forward and backward scanning SSMs.

In addition to the Time Mixer and Channel Mixer, learnable weighted average layers incorporate results from previous layers as described in equation \ref{eq:preliminaries_weighted_average}.

The \modelname{} encoder is a sequence to sequence model, meaning input and output dimension are equal. Optionally, a single learnable CLS token can be inserted before passing it through the encoder, meaning we would input and output a sequence of tokens of $\mathbb{R}^{d_{\texttt{model}} \times (L+1)}$. In section \ref{subsubsec:ablation_tokentype} we ablate different choices of CLS token types.

\subsubsection{Regression Head}\label{subsubsec:proposed_method_regression_head}
The regression head inputs the encoded sequence of tokens from \modelname{} encoder. 
If a CLS token is used, the regression head selects the the token representing the encoded CLS token and projects it from $\mathbb{R}^{d_{\texttt{model}}}$ into $\mathbb{R}$ using an MLP to obtain the final prediction of the state of health for a given cycle $k$. Note that the CLS could be at any position.

If no CLS token is used, we apply a mean operation to average the encoded sequence of tokens to obtain a single token representing the entire sequence. This token is then projected from $\mathbb{R}^{d_{\texttt{model}}}$ into $\mathbb{R}$ using an MLP to obtain the final prediction of the state of health for a given cycle $k$.

In section \ref{subsubsec:ablation_tokentype} we ablate different choices and positions of CLS token.

\subsection{Training}\label{subsec:proposed_method_training}
We train our \modelname{} model using the AdamW optimizer \citep{loshchilov_decoupled_2017} with a learning rate of $10^{-4}$, $\beta_1=0.9$ and $\beta_2=0.999$ and a weight decay of $5\cdot10^{-2}$. We use the mean squared error (MSE) loss function to train the model for 60 epochs. We use a step learning rate scheduler that halves the learning rate every 20 epochs. 
We randomly sample a batch of 32 discharge cycles of random batteries to predict the SOH of theses cycles. 

We apply drop-path regularization \citep{larsson_fractalnet_2016} with a drop-path rate of 0.2, where we occasionally drop entire mixer blocks. We further apply mixed precision training \citep{micikevicius_mixed_2017} to speed up the training.

During training, we use the our proposed anchor-based resampling technique to ensure that all cycles have the same number of samples while also acting as an augmentation technique. During sampling, we use linear resampling.

\subsection{Sampling}\label{subsec:proposed_method_sampling}

To recall, our \modelname{} model inputs a multi-variate time series of current, voltage, temperature and sample time of a single discharge cycle $k$ of a battery along with the time difference to the previous cycle $k-1$ and predicts the state of health $SOH_{k}$ of that cycle. We use the trained model to predict the SOH of a given cycle $k$ of a given battery $b_{\psi}$. To predict the complete capacity degradation of a battery, we iteratively predict the SOH of all cycles of a battery. 

In contrast to training, we use linear resampling to obtain time signals of the same length.

We highlight that in our sampling schema, the prediction of the SOH of a cycle $k$ is independent of the prediction of the SOH of the previous cycle $k-1$. This implies that the quality of the predictions is independent of the battery's history like number of cycles its has been charged and discharged and the profile of the discharge cycle. This choice is made to ensure that the model performs well in a realistic scenario where the battery's history is unknown.

\section{Experiments and Ablations}\label{sec:experiments}

In this section we present our results, experiments and ablations. We trained four different models of varying sizes as described in Table \ref{tab:experiment_models}.
\begin{table}[t]
    \centering
    \caption{Hyperparameters for our \modelname{} models of varying model size (for num\_samples = 128).}
    \label{tab:experiment_models}
    \begin{tabular}{ l || r r r r }
    \bf{Model} & $\mathbf{d_{model}}$ & $\mathbf{d_{state}}$ & \bf{\# layer} & \bf{\# Param} \\
    \hline
    \hline
    \modelname{}-S & 256 & 16 & 8 & 4.7\,M \\
    \modelname{}-M & 512 & 16 & 8 & 15.2\,M \\
    \modelname{}-L & 768 & 24 & 12 & 48.7\,M \\
    \modelname{}-XL & 1024 & 24 & 12 & 85.6\,M \\
    \end{tabular}
\end{table}

Throughout the experiments and ablations, we use \modelname{}-L trained on NASA-L (see Table \ref{tab:experiment_datasets}) as our base model if not explicitly stated otherwise.

\subsection{Dataset}\label{subsec:experiment_dataset}
We use the discharge cycles for a Li-ion Battery dataset from the NASA Ames Prognostics Center of Excellence (PCoE) \citep{saha2007nasa}.

As depicted in Table \ref{tab:experiment_nasa_battery_specs}, this dataset features multiple Li-ion batteries tested under various discharge profiles, ambient temperatures $T_{amb}$, cut-off voltages $V_{CO}$ and initial capacities.

\begin{table}[t]
    \centering
    \caption{Discharge specifications for various NASA Li-ion batteries. For the profile we report the discharge current signal form and the discharge amplitude. $T_{amb}$ is the ambient temperature, $V_{CO}$ is the cut-off voltage and Initial Capacity is the initial capacity of the battery at the beginning of the measurement campaign.}
    \label{tab:experiment_nasa_battery_specs}
    \begin{tabular}{ l || r r r r }
    \bf{ID} & \bf{Profile} & $\mathbf{T_{amb}}$ & $\mathbf{V_{CO}}$ & \bf{Initial Capacity} \\
    \hline
    \hline
    \#5 & (const.) 2.0A      & 24\,°C & 2.7\,V & 1.8565\,Ah \\
    \#6 & (const.) 2.0A      & 24\,°C & 2.5\,V & 2.0353\,Ah \\
    \#7 & (const.) 2.0A      & 24\,°C & 2.2\,V & 1.8911\,Ah \\
    \#18 & (const.) 2.0A     & 24\,°C & 2.5\,V & 1.8550\,Ah \\
    \#25 & (PWM 0.05Hz) 4.0A & 24\,°C & 2.0\,V & 1.8470\,Ah \\
    \#26 & (PWM 0.05Hz) 4.0A & 24\,°C & 2.2\,V & 1.8133\,Ah \\
    \#27 & (PWM 0.05Hz) 4.0A & 24\,°C & 2.5\,V & 1.8233\,Ah \\
    \#28 & (PWM 0.05Hz) 4.0A & 24\,°C & 2.7\,V & 1.8047\,Ah \\
    \#29 & (const.) 4.0A     & 43\,°C & 2.0\,V & 1.8447\,Ah \\
    \#31 & (const.) 1.5A     & 43\,°C & 2.5\,V & 1.8329\,Ah \\
    \#34 & (const.) 4.0A     & 24\,°C & 2.2\,V & 1.6623\,Ah \\
    \#36 & (const.) 2.0A     & 24\,°C & 2.7\,V & 1.8011\,Ah \\
    \#45 & (const.) 1.0A     &  4\,°C & 2.0\,V & 0.9280\,Ah \\
    \#46 & (const.) 1.0A     &  4\,°C & 2.2\,V & 1.5161\,Ah \\
    \#47 & (const.) 1.0A     &  4\,°C & 2.5\,V & 1.5244\,Ah \\
    \#48 & (const.) 1.0A     &  4\,°C & 2.7\,V & 1.5077\,Ah \\
    \#54 & (const.) 2.0A     &  4\,°C & 2.2\,V & 1.1665\,Ah \\
    \#55 & (const.) 2.0A     &  4\,°C & 2.5\,V & 1.3199\,Ah \\
    \#56 & (const.) 2.0A     &  4\,°C & 2.7\,V & 1.3444\,Ah \\

    \end{tabular}
\end{table}

All those batteries are 18650 NCA cells with a nominal capacity of 2000\,mAh and an upper voltage threshold of 4.2\,V.

In Table \ref{tab:experiment_datasets} we list various training and evaluation splits we compiled from those batteries. NASA-S is the same configuration \cite{mazzi_lithium-ion_2024} was using.

\begin{table}[t]
    \centering
    \caption{Different Training and Evaluation splits for the NASA Li-ion batteries used throughout our experiments and ablations.}
    \label{tab:experiment_datasets}
    \begin{tabular}{ l || r r r}
    \bf{ID} & \bf{NASA-S} & \bf{NASA-M} & \bf{NASA-L}\\
    \hline
    \hline
    \#5 & train & train & train\\
    \#6 & eval & eval & eval \\
    \#7 & eval & eval & eval \\
    \#18 & - & train & train \\
    \#25 & train & - & - \\
    \#26 & - & - & - \\
    \#27 & - & - & - \\
    \#28 & - & - & - \\
    \#29 & train & - & - \\
    \#31 & - & - & train \\
    \#34 & - & - & train \\
    \#36 & - & - & train \\
    \#45 & - & train & train \\
    \#46 & - & train & train \\
    \#47 & eval & eval & eval \\
    \#48 & train & train & train \\
    \#54 & - & - & train \\
    \#55 & - & - & train \\
    \#56 & - & - & train \\

    \end{tabular}
\end{table}

In our pre-processing, we remove cycles that have obvious issues with the measurement setup like those where the measured capacity drops occasionally to 0.0\,mAh. Explicitly we filter those cycles where from one cycle to the next the SOH drops more than 10\,\%. Further, for each cycle we remove those individual samples, that were recorded after the load has been disconnected. We also calculate the time between two cycles that we need for our positional encoding and we resample the time signals to have the same constant number of samples. During training we resample using our anchor-based resampling technique introduced in section \ref{subsec:proposed_method_architecture}. During inference we use linear resampling.

Throughout the experiments and ablations, we use NASA-L as our default dataset if not explicitly stated otherwise.

\begin{figure}[t]
    \centering
    \includegraphics[width=\linewidth]{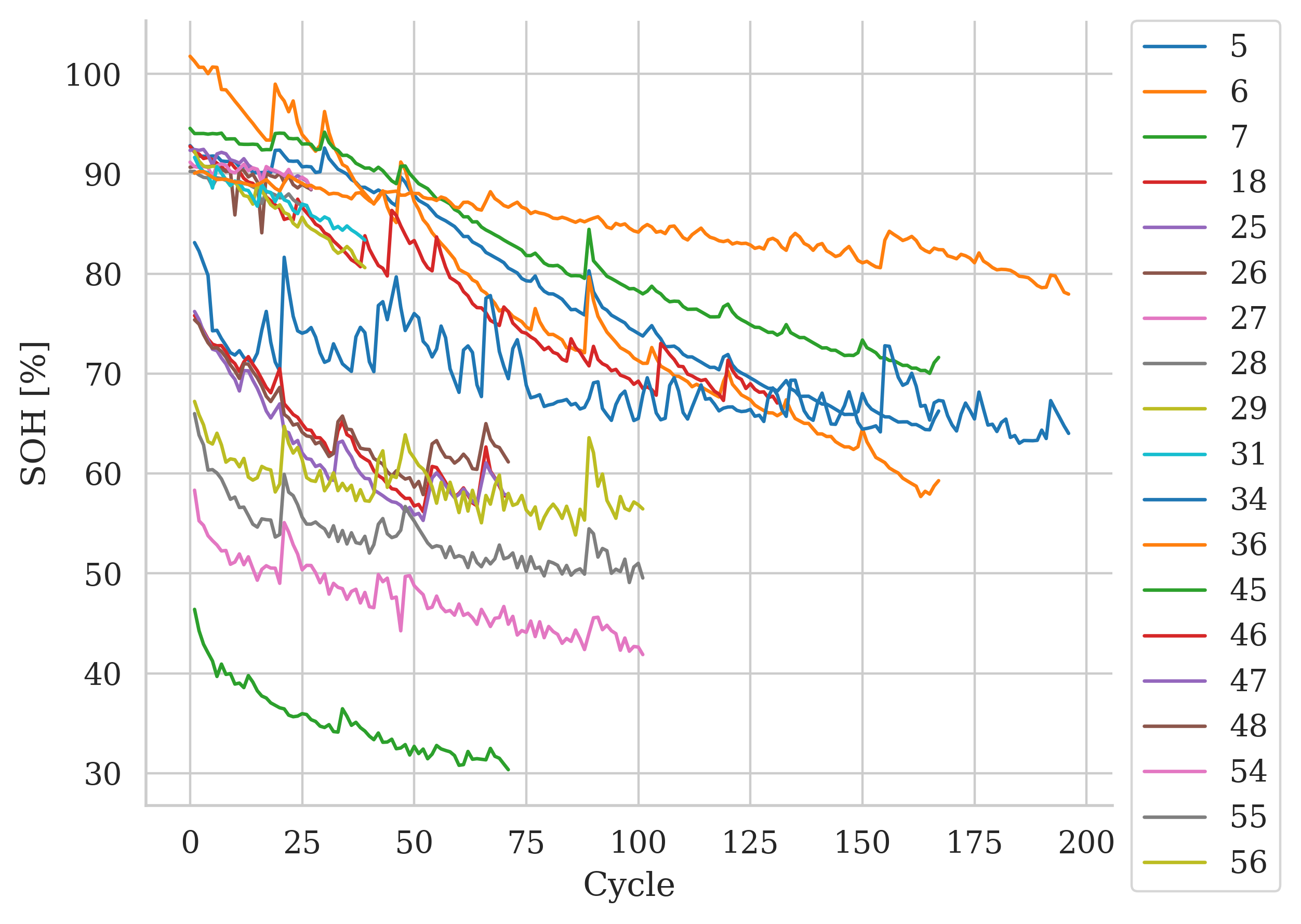}
    \caption{Capacity degradation for all selected batteries.}
    \label{fig:capacity_over_cycle}
\end{figure}

In Figure \ref{fig:capacity_over_cycle} we show the capacity degradation for all selected and pre-processed batteries. We illustrate the state of health (SOH) in percent over the discharge cycle ID.

\subsection{Metrics}\label{subsec:experiment_metrics}
We evaluate our experiments using the following commonly used metrics for state of health prediction tasks:

\begin{itemize}
    \item \textbf{MAE} mean absolute error:
    \begin{equation}
        \label{eq:mae}
        \text{MAE} = \frac{1}{K}\sum_{k=1}^{K}\left|\text{soh}^{\texttt{gt}}_k-\text{soh}^{\texttt{pred}}_k\right|.
    \end{equation}

    \item \textbf{RMSE} Root mean square error:
    \begin{equation}
        \label{eq:rmse}
        \text{RMSE} = \sqrt{\frac{1}{K}\sum_{k=1}^{K}\left(\text{soh}^{\texttt{gt}}_k-\text{soh}^{\texttt{pred}}_k\right)^2}.
    \end{equation}

    \item \textbf{MAPE} Mean Absolute Percentage Error:
    \begin{equation}
        \label{eq:mape}
        \text{MAPE} = \frac{1}{K}\sum_{k=1}^{K}\frac{\left|\text{soh}^{\texttt{gt}}_k-\text{soh}^{\texttt{pred}}_k\right|}{\left|\text{soh}^{\texttt{gt}}_k\right|},
    \end{equation}

    \item \textbf{AEOLE} Absolute End of Life Error:
    \begin{equation}
        \label{eq:aeole}
        \text{AEOLE} = \left| \text{eol}^{\texttt{gt}} - \text{eol}^{\texttt{pred}} \right|,
    \end{equation}
\end{itemize}

where $\text{soh}^{\texttt{gt}}_k$ is the ground truth for cycle $k$, $\text{soh}^{{\texttt{pred}}}_k$ is the predicted value for cycle $k$, $K$ is the total number of cycles, $\text{eol}^{\texttt{gt}}$ is the ground truth of the end of life indicator and $\text{eol}^{\texttt{pred}}$ is the prediction for the end of life indicator.

\subsection{Experiments}\label{subsec:experiment_experiments}

In this section we perform experiments with our \modelname{}-L model trained on NASA-L. In section \ref{subsubsec:experiment_entire_life_time} we show the SOH estimation for the entire battery lifetime. In section \ref{subsubsec:experiment_dataset_split} we show the performance of our model when trained on differently sized datasets. In section \ref{subsubsec:experiment_model_scaling} we show the performance of our model when scaling the model size as well the dataset size. In section \ref{subsubsec:experiment_soh_estimation_different_starting_points} we show the performance of our model when starting the prediction at different cycle IDs simulating pre-aged batteries.

\subsubsection{SOH Estimation for Entire Battery Lifetime}\label{subsubsec:experiment_entire_life_time}

\begin{figure}[t]
    \includegraphics[width=\linewidth]{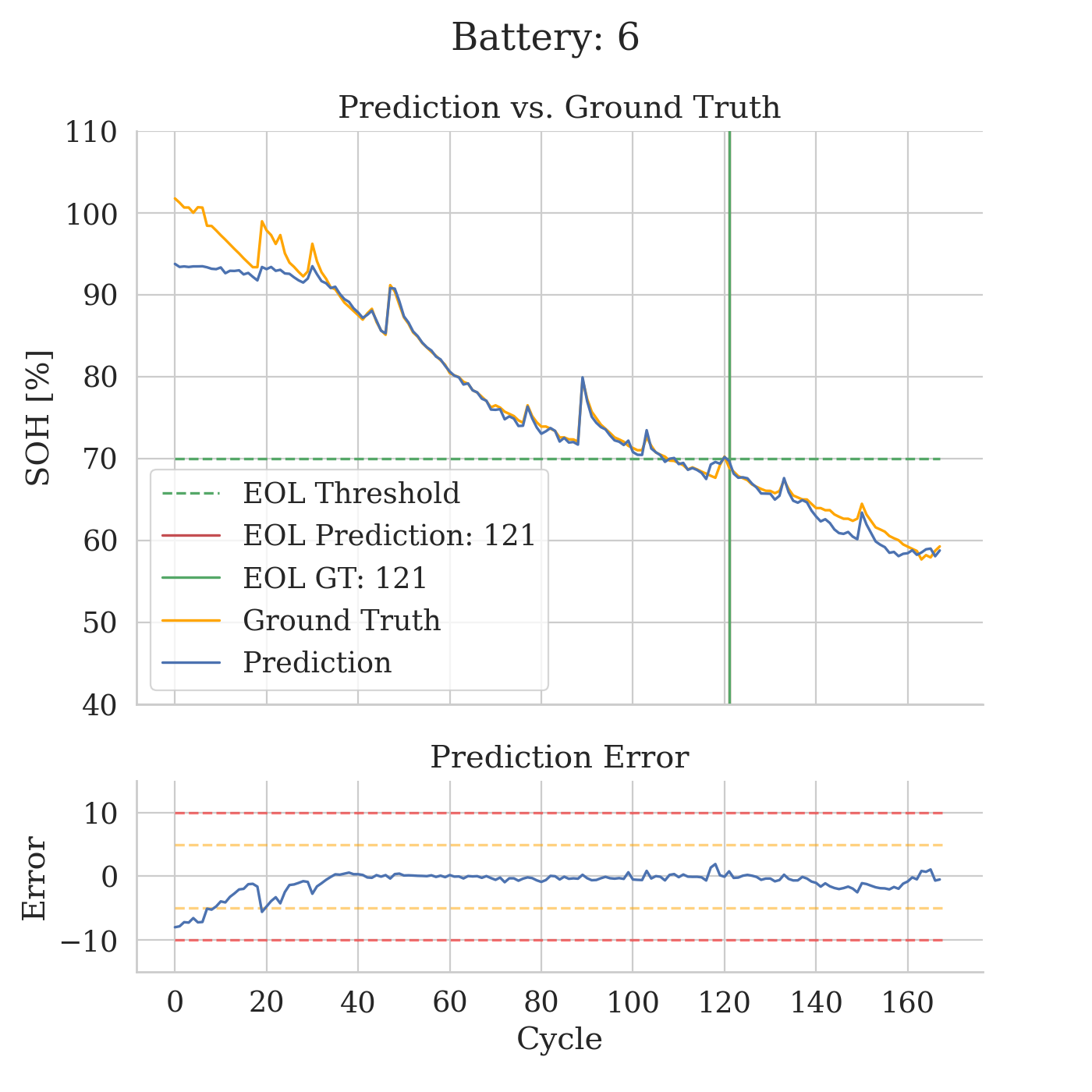}
    \caption{SOH prediction for Battery \#06}
    \label{fig:soh_prediction_bat6}
\end{figure}

\begin{figure}[t]
    \includegraphics[width=\linewidth]{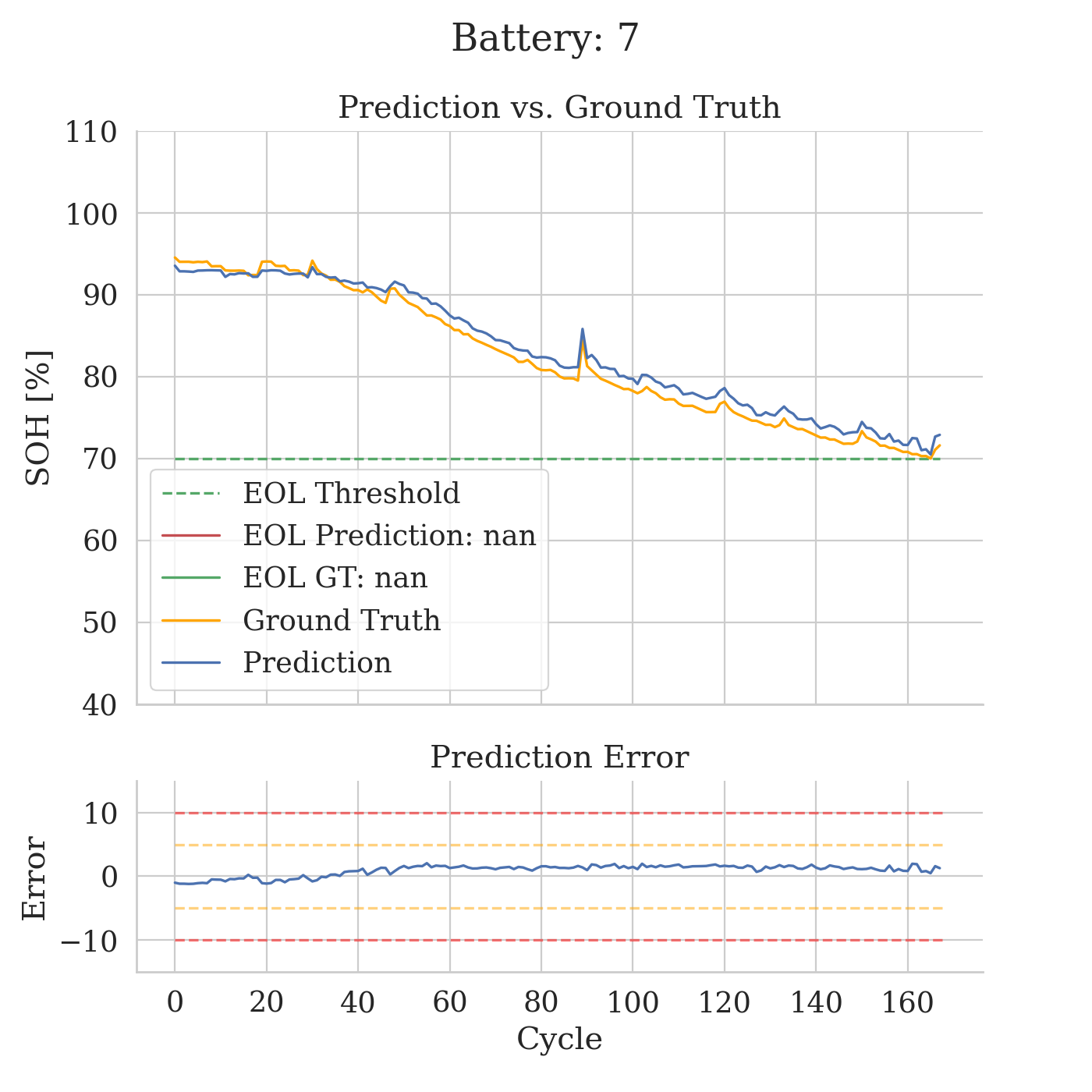}
    \caption{SOH prediction for Battery \#07}
    \label{fig:soh_prediction_bat7}
\end{figure}

\begin{figure}[t]
    \includegraphics[width=\linewidth]{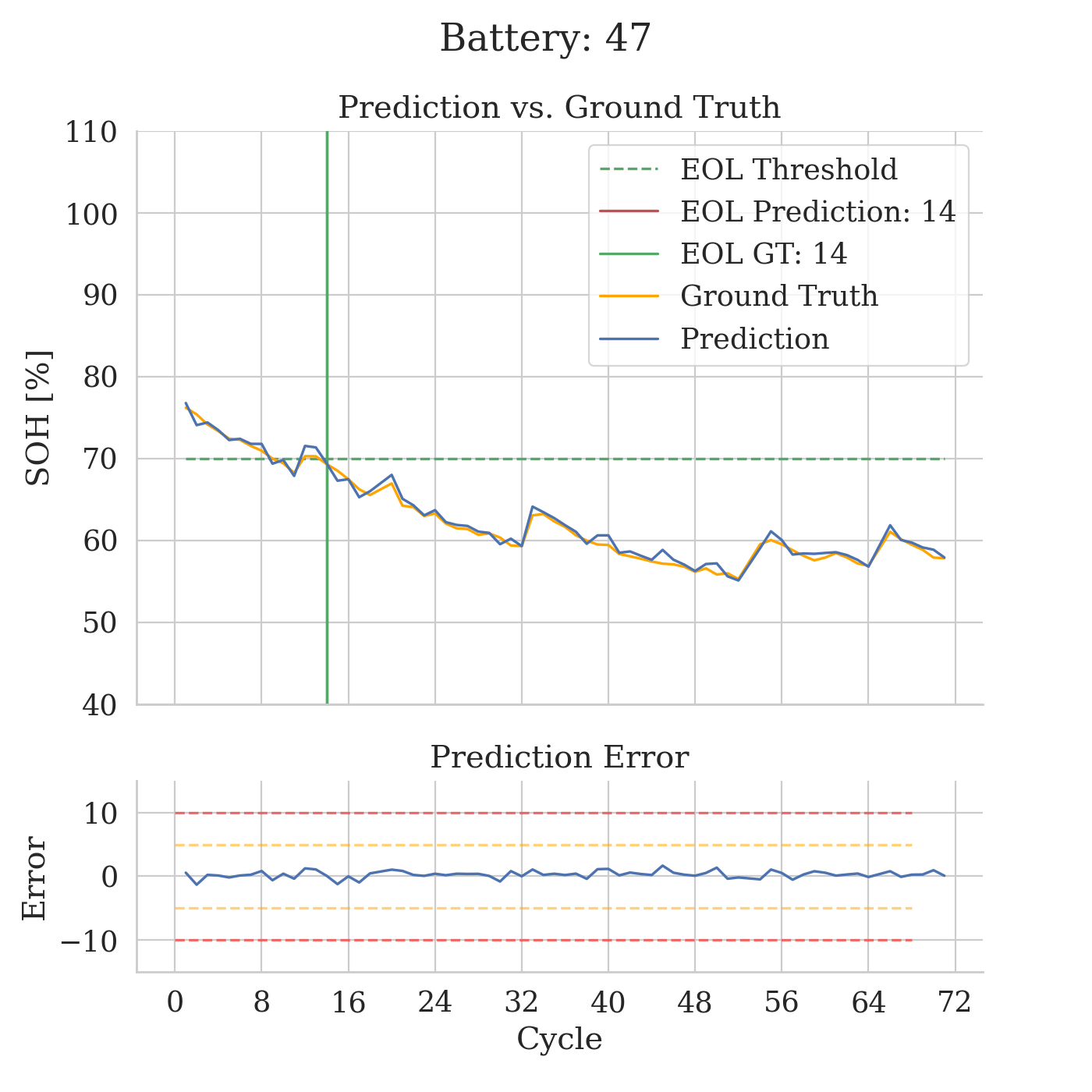}
    \caption{SOH prediction for Battery \#47}
    \label{fig:soh_prediction_bat47}
\end{figure}

As described in section \ref{sec:proposed_method}, we input the resampled time signal from a single discharge cycle and predict the state of health of the battery for that particular cycle. If we sample the model as described in section \ref{subsec:proposed_method_sampling} we can obtain the capacity degradation over the cycle ID for each battery in the evaluation set. Figures \ref{fig:soh_prediction_bat6}, \ref{fig:soh_prediction_bat7}, \ref{fig:soh_prediction_bat47} depict the comparison of the predicted SOH values against the ground truth SOH values. We further show the error for each cycle as well as the resulting EOL indicator. 

The EOL indicator predicts at which cycle the battery reaches its end of life. It is defined as the first cycle bellow the EOL threshold. Due to recuperation effects of Li-ion batteries it is important to consider the last occurrence where the SOH value drops bellow the EOL threshold.

\begin{figure}[t]
    \includegraphics[width=\linewidth]{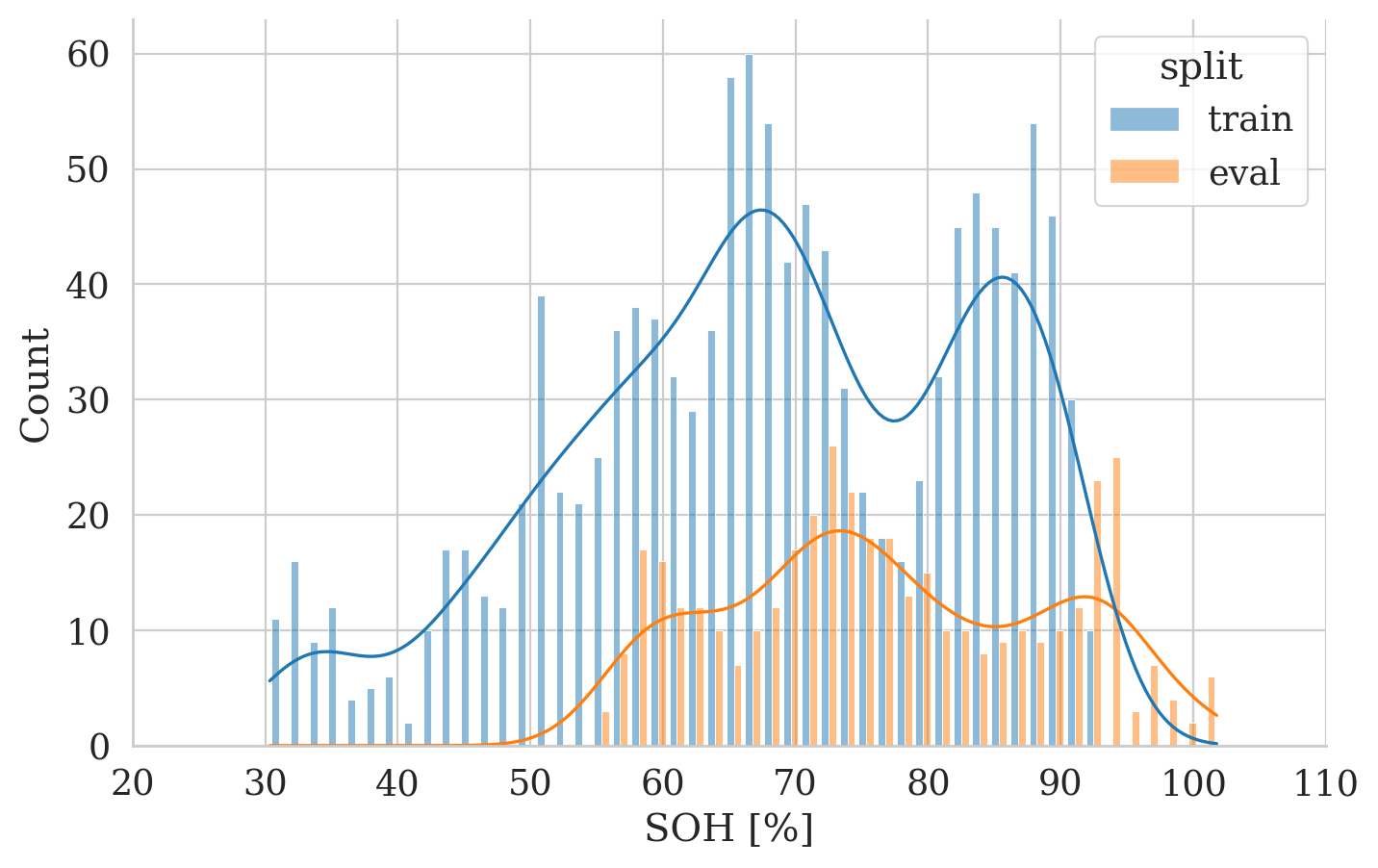}
    \caption{Histogram of SOH value counts. Comparison of train and eval split of the NASA-L dataset. Number of bins: 50.}
    \label{fig:mazzi_L_distribution}
\end{figure}

We observe that for the evaluation batteries \#06, \#07 and \#47 our \modelname{} model accurately predicts the dynamics of the SOH curves and predicts the EOL indicator without error. We notice that for battery \#06 the prediction for SOH values above 92\,\% has a comparably large error. We hypothesize that the model does not generalize well given the fact that the dataset is relatively small and that the training set does not contain samples with SOH values above 92\,\% (see Fig. \ref{fig:mazzi_L_distribution}).

Further, other Mamba-like models such as \cite{li_videomamba_2024} and \cite{ liu_vmamba_2024} have had similar issues with models overfitting easily.

In Table \ref{tab:experiments_results} we compare our \modelname{} model against \cite{mazzi_lithium-ion_2024} for each battery of the evaluation set.

\begin{table}[t]
    \centering
    \caption{Comparing our \modelname{} models with the state-of-the-art \cite{mazzi_lithium-ion_2024} on the NASA Li-ion batteries. We report the MAE, RMSE and MAPE for each battery. The best results are highlighted in bold.}
    \label{tab:experiments_results}
    \begin{tabular}{ l || l | r r r }
    \bf{Battery} & \bf{Model} & \bf{MAE}$\downarrow$ & \bf{RMSE}$\downarrow$ & \bf{MAPE}$\downarrow$ \\
    \hline
    \hline
    \#06 & \citeauthor{mazzi_lithium-ion_2024} & 2.448      & 3.177      & 1.579 \\
         & \modelname{} (ours)                 & \bf{1.173} & \bf{2.068} & \bf{1.406} \\
    \hline
    \#07 & \citeauthor{mazzi_lithium-ion_2024} & 1.861      & 2.252      & 1.114 \\
         & \modelname{} (ours)                 & \bf{1.197} & \bf{1.285} & \bf{1.498} \\
    \hline
    \#47 & \citeauthor{mazzi_lithium-ion_2024} & 2.549      & 3.094      & 1.969 \\
         & \modelname{} (ours)                 & \bf{0.512} & \bf{0.645} & \bf{0.822}
    \end{tabular}
\end{table}

We observe that our \modelname{} model surpasses \cite{mazzi_lithium-ion_2024} in all metrics for all batteries. Later in section \ref{tab:experiments_model_scaling} we show how our method compares against \cite{mazzi_lithium-ion_2024} for different model sizes and and datasets.

\subsubsection{Dataset Split}\label{subsubsec:experiment_dataset_split}

In this experiment we test the performance of our \modelname{} model when trained on different training sets and compere those results against \cite{mazzi_lithium-ion_2024}. Explicitly, we train our \modelname{}-L model on NASA-S, NASA-M and NASA-L. Results are reported in Table \ref{tab:experiments_data_split}.

\begin{table}[t]
    \centering
    \caption{Performance of our \modelname{} model when trained on different training sets. Evaluation sets are the same for all datasets.}
    \label{tab:experiments_data_split}
    \begin{tabular}{ l l || r r r r }
    \bf{Model} & \bf{Dataset} & \bf{MAE}$\downarrow$ & \bf{RMSE}$\downarrow$ & \bf{MAPE}$\downarrow$ \\
    \hline
    \hline
    \citeauthor{mazzi_lithium-ion_2024} & NASA-S & 2.220 & 2.778 & 1.451 \\
    \hline
    \modelname{} (ours)  & NASA-S & 1.764 & 2.404 & 2.320  \\
                        & NASA-M & 1.334 & 1.902 & 1.641 \\
                        & NASA-L & \bf{1.072} & \bf{1.592} & \bf{1.346}
    \end{tabular}
\end{table}

We observe that our \modelname{} model performs better on MAE and RMSE for all datasets and performs better at MAPE for NASA-L.

\subsubsection{Model Scaling}\label{subsubsec:experiment_model_scaling}

In this experiment we test the performance of our \modelname{} model when trained with differently sized models. We train our \modelname{}-S, \modelname{}-M, \modelname{}-L and \modelname{}-XL models on NASA-S, NASA-M and NASA-L. The results are reported in Table \ref{tab:experiments_model_scaling}.

\begin{table}[t]
    \centering
    \caption{Model scaling experiment. We report the metrics MAE, RMSE and MAPE for the SOH estimation task for different model sizes and datasets.}
    \label{tab:experiments_model_scaling}
    \begin{tabular}{ l l || r r r r }
    \bf{Model} & \bf{Dataset} & \bf{MAE}$\downarrow$ & \bf{RMSE}$\downarrow$ & \bf{MAPE}$\downarrow$ \\
    \hline
    \hline
    \modelname{}-S & NASA-S & 2.478 & 3.974 & 3.325 \\
                   & NASA-M & 1.920 & 2.829 & 2.461 \\
                   & NASA-L & 1.895 & 2.929 & 2.315 \\
    \hline
    \modelname{}-M & NASA-S  & 1.987 & 2.879 & 2.609\\
                    & NASA-M & 1.736 & 2.414 & 2.170 \\
                    & NASA-L & 1.230 & 2.027 & 1.493 \\
    \hline
    \modelname{}-L & NASA-S & 1.764 & 2.404 & 2.320  \\
                   & NASA-M & 1.334 & 1.902 & 1.641 \\
                   & NASA-L & \bf{1.072} & \bf{1.592} & \bf{1.346} \\
    \hline
    \modelname{}-XL & NASA-S & 1.693 & 2.431 & 2.218 \\
                    & NASA-M & 1.349 & 1.966 & 1.642 \\
                    & NASA-L & 1.133 & 1.800 & 1.396 \\
    \end{tabular}
\end{table}

We can see that the performance of our model increases with the model size and the size of the dataset. This is expected since larger models have more capacity to learn complex patterns in the data and larger datasets provide more data for the model to learn from.

\begin{figure}[t]
    \includegraphics[width=\linewidth]{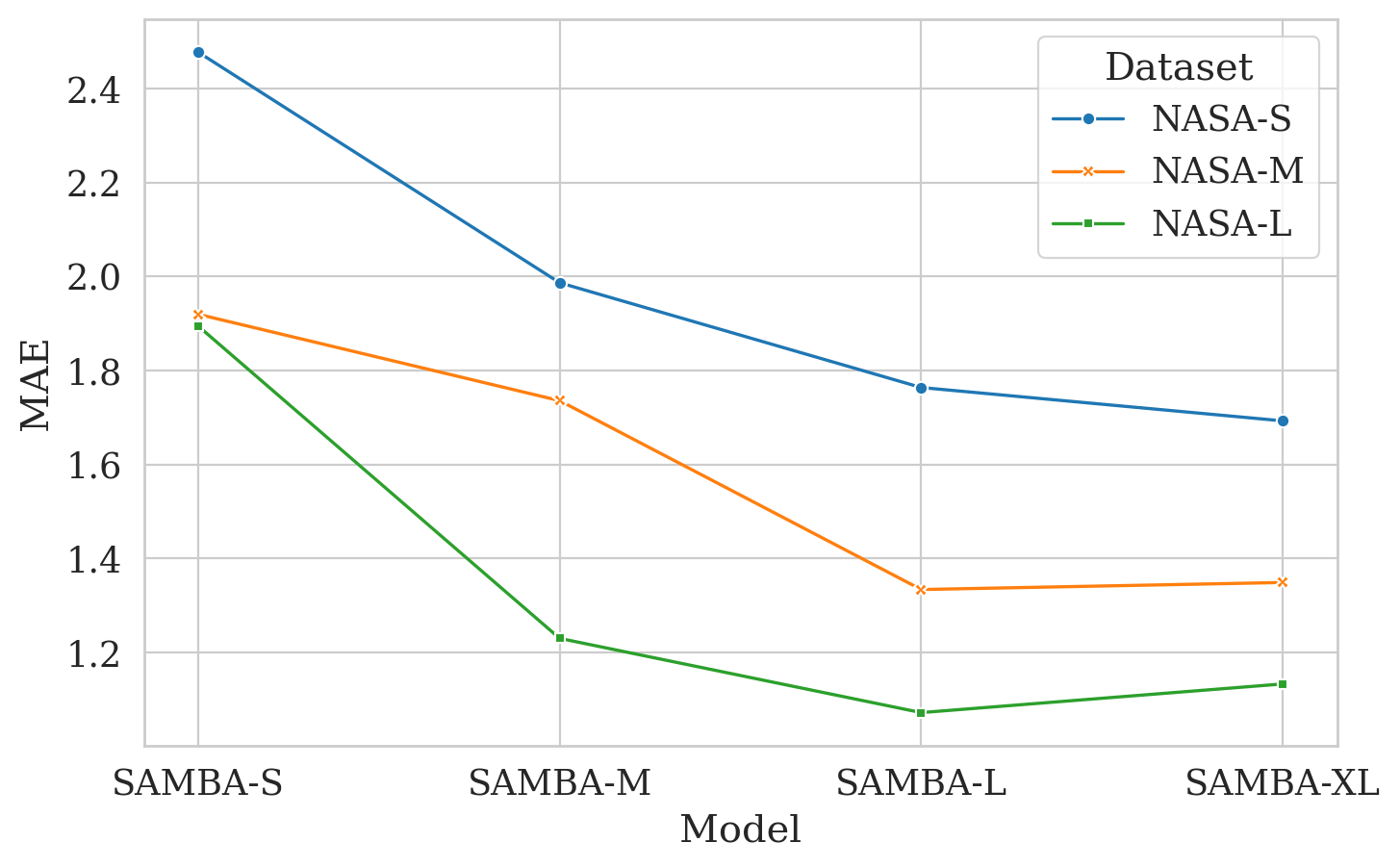}
    \caption{Model scaling experiment. MAE metric for the SOH estimation task for different model sizes and datasets. Values are reported in Table \ref{tab:experiments_model_scaling}}
    \label{fig:model_scaling}
\end{figure}

Figure \ref{fig:model_scaling} plots the MAE for the SOH estimation task for the different model sizes and datasets. We can observe that that for \modelname{}-S increasing the dataset size from NASA-M to NASA-L has almost no impact on the performance, indicating that the model is too small to learn from the additional data. 
Further, increasing the model size from \modelname{}-L to \modelname{}-XL decreases the performance slightly indicating that the model is too large for the dataset and likely overfits to the training data. 

\subsubsection{SOH Estimation for Used Batteries}\label{subsubsec:experiment_soh_estimation_different_starting_points}

In a real scenario, one will likely not always need to predict the SOH for new batteries, but also for batteries that have been used for an unknown number of cycles or probably not all discharge cycles have been recorded. A robust model is expected to still reliably predict the SOH values for such scenarios.

To simulate the prediction task of used batteries, we take the batteries from the evaluation set, remove the first discharge cycles and update their cycle ID. Explicitly, for batteries \#06 and \#07 we experiment starting the prediction at cycle 0, 30, 70 and 100 and for battery \#47 with 0, 15, 35 and 50. In Table \ref{tab:experiments_different_starting_points} we report our results.

\begin{table}[t]
    \centering
    \caption{SOH estimation performance on the evaluation batteries starting at different cycle IDs. We report the metrics MAE, RMSE and MAPE for the SOH estimation task and the AEOLE for EOL indication. Capital letters in brackets for the start column represent \citeauthor{mazzi_lithium-ion_2024} notation for those scenarios. N/R=Not Reported.}
    \label{tab:experiments_different_starting_points}
    \begin{tabular}{ l r || r r r r }
    \bf{Model} & \bf{Start} & \bf{MAE}$\downarrow$ & \bf{RMSE}$\downarrow$ & \bf{MAPE}$\downarrow$ & \bf{AEOLE}$\downarrow$ \\
    \hline
    \hline
    \bf{Battery \#06} & & & & & \\
    \hline
    \citeauthor{mazzi_lithium-ion_2024} &       0 & 2.448 & 3.177 & 1.579 & N/R \\
                                        &  30 (A) & 2.445 & 3.090 & 1.726 & \bf{0} \\
                                        &  70 (C) & 2.080 & 2.516 & 1.650 & 3 \\
                                        & 100 (E) & 2.440 & 2.859 & 1.901 & \bf{0} \\
    \hline
    \modelname{}                        &       0 & \bf{1.173} & \bf{2.068} & \bf{1.406} & \bf{0} \\
                                        &  30 (A) & \bf{0.575} & \bf{0.824} & \bf{0.845} & \bf{0} \\
                                        &  70 (C) & \bf{0.680} & \bf{0.905} & \bf{1.045} & \bf{0} \\
                                        & 100 (E) & \bf{0.808} & \bf{1.045} & \bf{1.275} & \bf{0} \\
    \hline
    \bf{Battery \#07} & & & & & \\
    \hline
    \citeauthor{mazzi_lithium-ion_2024} &       0 & 1.861 & 2.252 & \bf{1.114} & N/R \\
                                        &  30 (B) & 1.748 & 2.285 & \bf{1.092} & N/R \\
                                        &  70 (D) & 1.794 & 2.101 & \bf{1.180} & N/R \\
                                        & 100 (F) & 1.608 & 1.868 & \bf{1.011} & N/R \\
    \hline
    \modelname{}                        &       0 & \bf{1.197} & \bf{1.285} & 1.498 & \bf{0} \\
                                        &  30 (B) & \bf{1.309} & \bf{1.371} & 1.665 & \bf{0} \\
                                        &  70 (D) & \bf{1.400} & \bf{1.433} & 1.839 & \bf{0} \\
                                        & 100 (F) & \bf{1.395} & \bf{1.434} & 1.878 & \bf{0} \\
    \hline
    \bf{Battery \#47} & & & & & \\
    \hline
    \citeauthor{mazzi_lithium-ion_2024} &       0 & 2.549 & 3.094 & 1.969 & N/R \\
                                        &  15 (G) & 2.774 & 3.491 & 2.345 & N/R \\
                                        &  35 (H) & 2.110 & 2.540 & 1.841 & N/R \\
                                        &  50 (I) & 1.806 & 2.416 & 1.570 & N/R \\
    \hline
    \modelname{}                        &       0 & \bf{0.512} & \bf{0.645} & \bf{0.822} & \bf{0} \\
                                        &  15 (G) & \bf{0.507} & \bf{0.638} & \bf{0.843} & \bf{0} \\
                                        &  35 (H) & \bf{0.508} & \bf{0.638} & \bf{0.871} & \bf{0} \\
                                        &  50 (I) & \bf{0.480} & \bf{0.592} & \bf{0.825} & \bf{0}
    
    \end{tabular}
\end{table}

We observe that \modelname{} performs better on all reported metrics for all batteries and starting points, except the MAPE for battery \#07. Since our \modelname{} model performs the prediction task independently for each cycle individually, our method is robust against missing cycles and batteries of different age. The SOH prediction curve is exactly the same. The metrics only vary for different starting points since the metrics are normalized by the total number of cycles $K$ for each battery.

\subsection{Ablation Study}\label{subsec:ablation_study}
In this section we ablate our contributions and design choices. If not stated otherwise, we use our \modelname{}-L model trained on NASA-L. In section \ref{subsubsec:ablation_tokentype} we ablate the usage and position of the class tokens that can optionally be inserted into the input token sequence. In section \ref{subsubsec:ablation_backbone} we ablate the performance of our \modelname{} backbone and compare it with a vanilla Mamba backbone from \citep{gu_mamba_2024}. We continue investigating the performance for various resampling techniques in section \ref{subsubsec:ablation_resampling}. Finally, we test the performance for different input projections and position encodings in section \ref{subsubsec:ablation_positional_encoding}.

\subsubsection{Usage and Position of Class Token}\label{subsubsec:ablation_tokentype}
We ablate the usage and the potential position of class tokens inserted into the token sequence. We train our \modelname{}-L model on NASA-L inserting a class token either at the tail, middle or head and compare it with a model that inserts no class token. If we use a class token, the head is attached to the position at the output that corresponds to the position where the class token was placed. If no class token is used, we average the output of all output tokens and feed it to the regression head. The results are reported in Table \ref{tab:ablation_cls_token_type}. 

\begin{table}[t]
    \centering
    \caption{Ablation of inserting a class token into the input token sequence and at which positions.}
    \label{tab:ablation_cls_token_type}
    \begin{tabular}{ l || r r r r}
    \bf{CLS Token Type} & \bf{MAE}$\downarrow$ & \bf{RMSE}$\downarrow$ & \bf{MAPE}$\downarrow$ \\
    \hline
    \hline
    Tail & 5.515 & 8.141 & 6.612  \\
    Middle & 1.977 & 4.131 & 2.260  \\
    Head & 1.746 & 3.384 & 2.029 \\
    None (Avg.) & \bf{1.072} & \bf{1.592} & \bf{1.346} \\
    \end{tabular}
\end{table}

\subsubsection{Backbone}\label{subsubsec:ablation_backbone}
In this ablation we compare the performance of our \modelname{} backbone with the vanilla Mamba backbone from \cite{gu_mamba_2024}. We train both models on NASA-L. The results are shown in Table \ref{tab:ablation_backbone}. The main motivation of this ablation is to show the effectiveness of our \modelname{} backbone when it comes to multi-variate time signals.

\begin{table}[t]
    \centering
    \caption{Ablation of different backbone architectures.}
    \label{tab:ablation_backbone}
    \begin{tabular}{ l || r r r}
    \bf{Backbone} & \bf{MAE}$\downarrow$ & \bf{RMSE}$\downarrow$ & \bf{MAPE}$\downarrow$ \\
    \hline
    \hline
    Vanilla Mamba & 1.709 & 2.386 & 2.161 \\
    \modelname{} (ours) & \bf{1.072} & \bf{1.592} & \bf{1.346} \\
    \end{tabular}
\end{table}

We can see that our \modelname{} backbone outperforms the vanilla Mamba backbone. This is due to the fact that the \modelname{} backbone is designed to handle multi-variate time signals and is able to capture the complex relationships between the different variables in the dataset.

\subsubsection{Resampling}\label{subsubsec:ablation_resampling}
In this ablation we compare the performance of different resampling methods. We train our \modelname{}-L model on NASA-L using linear, random and our proposed anchor-based resampling. The results are shown in Table~\ref{tab:ablation_resampling}. The target of this ablation is to show the effectiveness of our anchor-based resampling method introduced in section \ref{subsubsec:proposed_method_anchor_based_resampling}.

\begin{table}[t]
    \centering
    \caption{Ablation of various resampling methods.}
    \label{tab:ablation_resampling}
    \begin{tabular}{ l || r r r}
    \bf{Resample Type} & \bf{MAE}$\downarrow$ & \bf{RMSE}$\downarrow$ & \bf{MAPE}$\downarrow$ \\
    \hline
    \hline
    Linear & 1.272 & 1.862 & 1.631 \\
    Random & 3.315 & 4.368 & 4.302 \\
    Anchors (ours) & \bf{1.072} & \bf{1.592} & \bf{1.346} \\
    \end{tabular}
\end{table}

Our anchor-based resampling method outperforms the linear and random resampling methods. We hypothesize that this is due to the fact that the anchor-based resampling acts as a form of data augmentation, allowing the model to learn more robust features from the data. 

\subsubsection{Positional Encoding}\label{subsubsec:ablation_positional_encoding}
In this ablation we compare the performance of different positional encoding methods to justify our choice of the sample time positional encoding introduced in section \ref{subsubsec:proposed_method_sample_time_position_embeddings}. We train our \modelname{}-L model on NASA-L using no encoding, sample time encoding and our proposed combined sample time and cycle time difference encoding. The results are shown in Table~\ref{tab:ablation_positional_encoding}. 

\begin{table}[t]
    \centering
    \caption{Ablation for various positional encoding methods.}
    \label{tab:ablation_positional_encoding}
    \begin{tabular}{ l || r r r}
        \bf{Encoding Type} & \bf{MAE}$\downarrow$ & \bf{RMSE}$\downarrow$ & \bf{MAPE}$\downarrow$ \\
        \hline
        \hline
        No Encoding & 3.097 & 3.966 & 4.257 \\
        Sample Time & 1.160 & 1.721 & 1.450 \\
        Sample Time + Cycle Diff (ours) & \bf{1.072} & \bf{1.592} & \bf{1.346} \\
    \end{tabular}
\end{table}

Clearly, adding our proposed positional encoding to the model improves the performance. Further adding the time difference between discharge cycles as an additional feature to the positional encoding increases the performance even further. The intuition is that the difference between discharge cycles is important to capture recuperation effects of the battery and adjust the prediction accordingly.

\section{Conclusion}\label{sec:conclusion}
We have presented \modelname{}, a novel approach for the prediction of the state of health of Li-ion batteries on structured state space model. We have shown that our model outperforms the state-of-the-art on the NASA battery discharge dataset~\cite{saha2007nasa}. We further introduced a novel anchor-based resampling method and a sample time and cycle time difference positional encoding to improve the performance of our model. Our results show that our model is able to predict the state of health of Li-ion batteries with high accuracy and robustness, capable to extract information from multi-variate time series data and to model recuperation effects.

\subsection{Limitations}\label{subsec:limitations}
Even though our model outperforms the state-of-the-art on the NASA battery discharge dataset, we acknowledge that we evaluated our model only on a single dataset; the NASA battery discharge dataset from~\cite{saha2007nasa}. This dataset only contains batteries of the same chemistry and we selected only constant discharge cycles for our experiments. Future work should evaluate our model on different datasets and different battery chemistries to further validate the generalization capabilities of our method.

\subsection{Future Work}\label{subsec:future_work}
In future work, we plan to evaluate our model on different datasets and different battery chemistries to further validate the generalization capabilities of our model. We also plan to investigate the impact of different discharge profiles on the performance of our model. Furthermore, we plan to investigate the impact of different hyperparameters on the performance of our model and to further optimize our model for better performance. Finally, we plan to investigate different model architectures and different state space models to further improve the performance of our model.

\section*{Acknowledgements}
This publication is part of the In4Labs project with reference TED2021-131535BI00 funded by MICIU/AEI/10.13039/501100011033 and by “European Union Next Generation EU/PRTR”. 

\bibliographystyle{plainnat}
\bibliography{main}


\newpage 

\setlength\intextsep{0pt} 
\begin{wrapfigure}{l}{25mm}
    \includegraphics[width=1in,height=1.25in,clip,keepaspectratio]{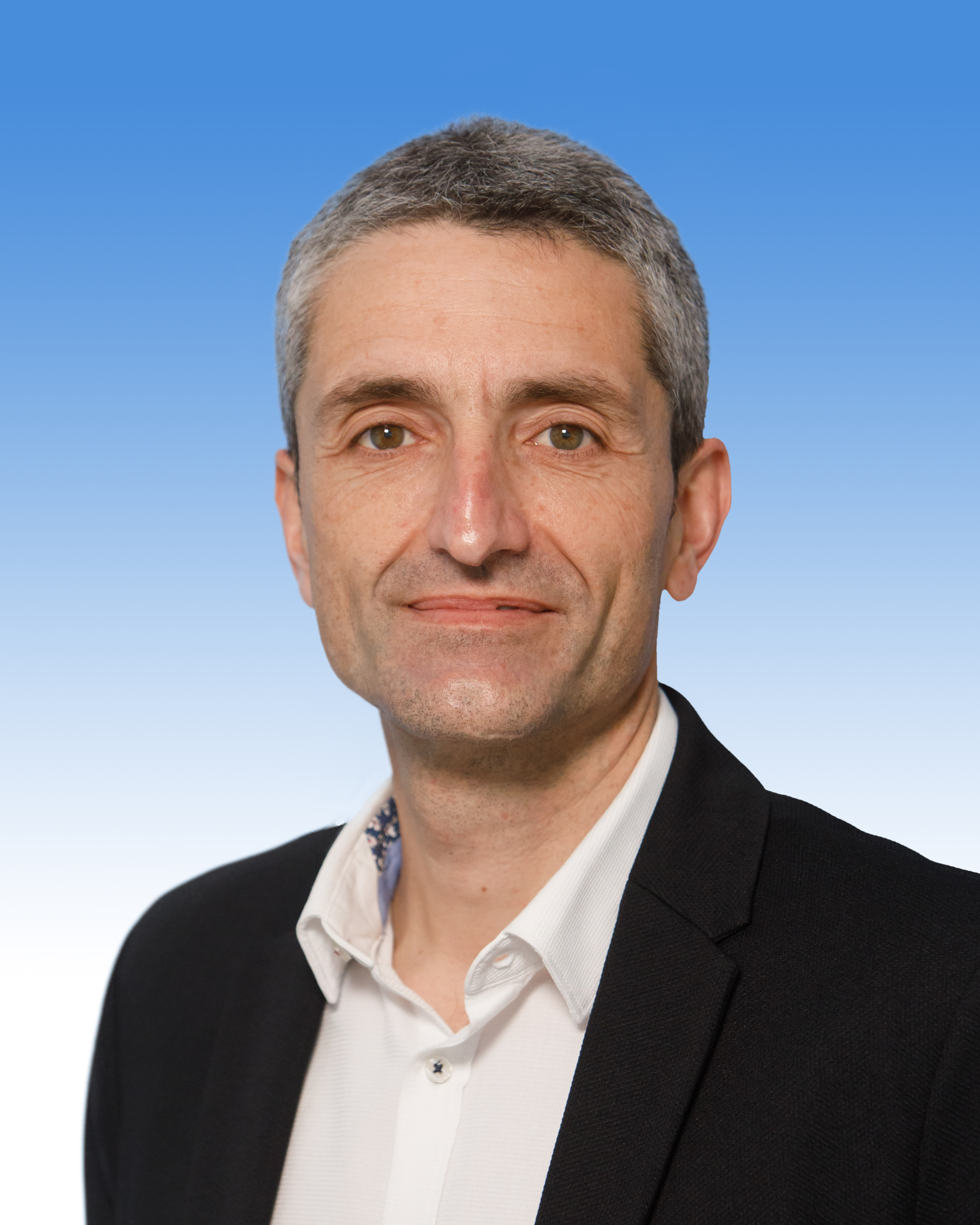}
\end{wrapfigure}\par
\noindent \textbf{José Ignacio Olalde-Verano} is a doctoral student at UNED, Spain. His research focuses on machine learning techniques applied to industry 4.0. Master's Degree in Research in Industrial Technologies at UNED, studied Technical Engineering at the University of Zaragoza and adapted to the degree at the University of León. Since 2009 works in the automotive industry.\\

\setlength\intextsep{0pt} 
\begin{wrapfigure}{l}{25mm}
    \includegraphics[width=1in,height=1.25in,clip,keepaspectratio]{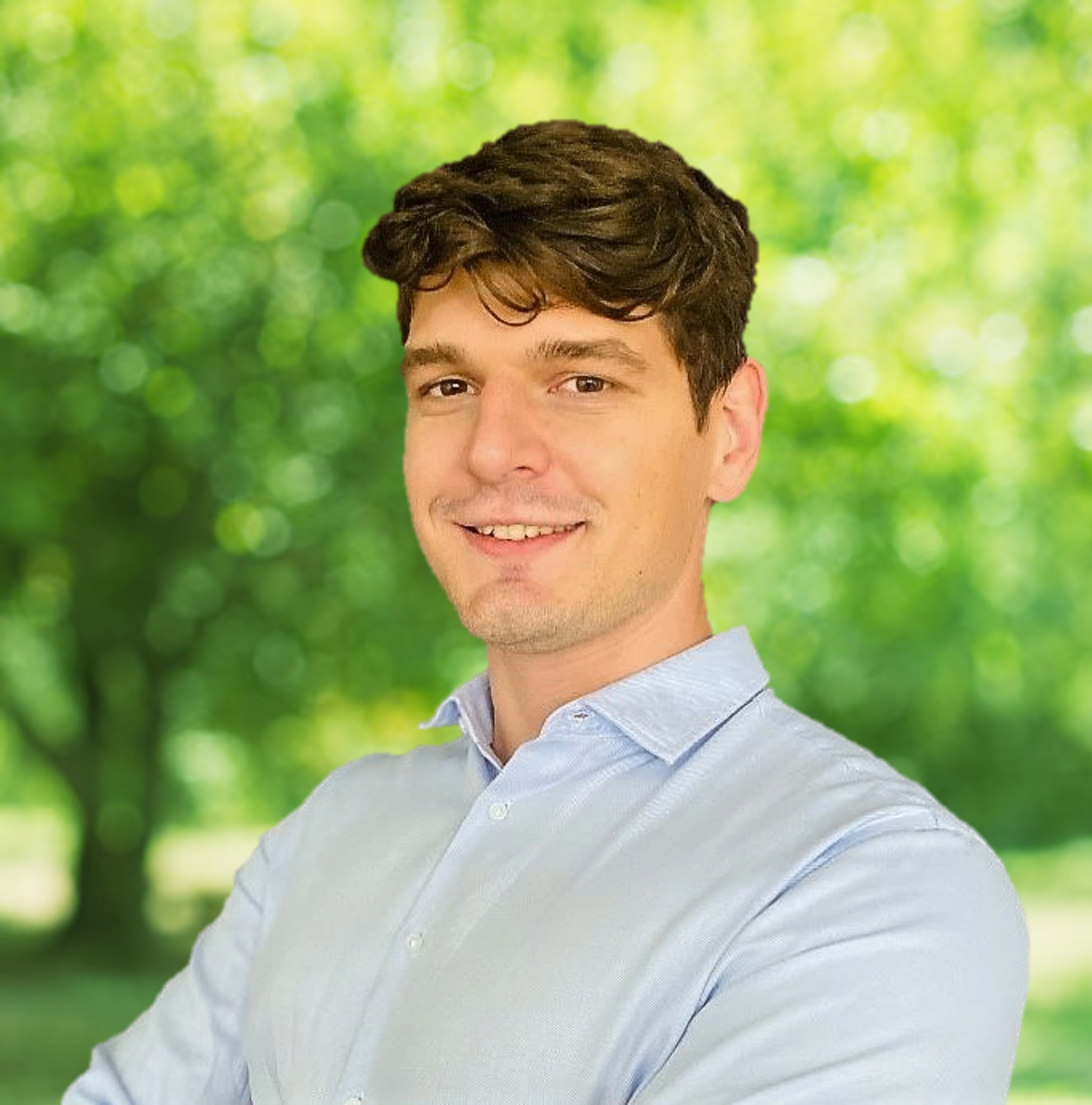}
\end{wrapfigure}\par
\noindent \textbf{Sascha Kirch} is a doctoral student at UNED, Spain. His research focuses on self-supervised multi-modal generative deeplearning. He received his M.Sc. degree in Electronic Systems for Communication and Information from UNED, Spain. He received his B.Eng. degree in electrical engineering from the Cooperative State University Baden-Wuerttemberg (DHBW), Germany. Sascha is member of IEEE’s honor society Eta Kappa Nu and president of the chapter Nu Alpha.\\

\setlength\intextsep{0pt} 
\begin{wrapfigure}{l}{25mm}
    \includegraphics[width=1in,height=1.25in,clip,keepaspectratio]{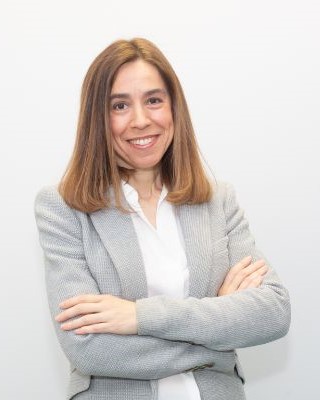}
\end{wrapfigure}\par
\noindent \textbf{Clara Pérez-Molina} received her M.Sc. degree in Physics from the Complutense University in Madrid and her PhD in Industrial Engineering from the Spanish University for Distance Education (UNED). She has worked as researcher in several National and European Projects and has published different technical reports and research articles for International Journals and Conferences, as well as several teaching books. She is currently an Associate Professor with tenure of the Electrical and Computer Engineering Department at UNED.
Her research activities are centered on Educational Competences and Technology Enhanced Learning applied to Higher Education in addition to Renewable Energy Management and Artificial Intelligence techniques.\\

\setlength\intextsep{0pt} 
\begin{wrapfigure}{l}{25mm}
    \includegraphics[width=1in,height=1.25in,clip,keepaspectratio]{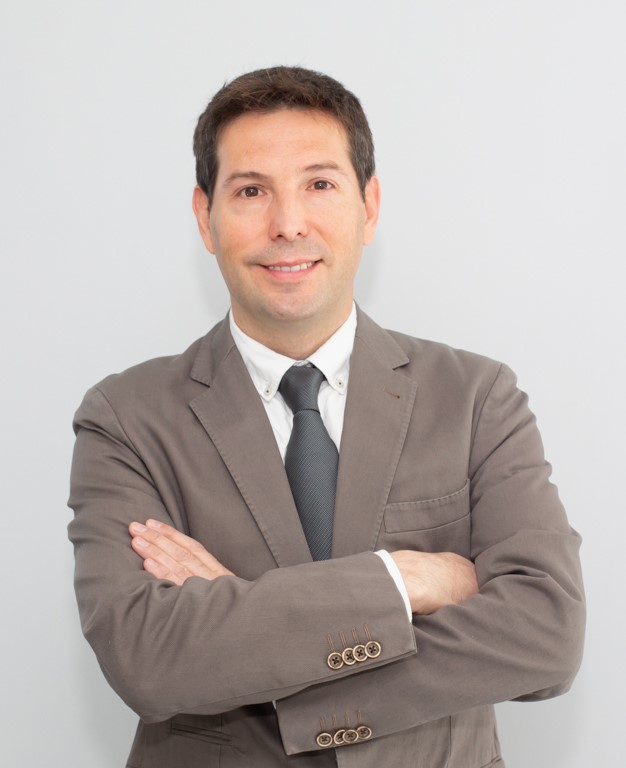}
\end{wrapfigure}\par
\noindent \textbf{Sergio Martín} is Associate Professor at UNED (National University for Distance Education, Spain). He is PhD by the Electrical and Computer Engineering Department of the Industrial Engineering School of UNED. He is Computer Engineer in Distributed Applications and Systems by the Carlos III University of Madrid. He teaches subjects related to microelectronics and digital electronics since 2007 in the Industrial Engineering School of UNED. He has participated since 2002 in national and international research projects related to mobile devices, ambient intelligence, and location-based technologies as well as in projects related to "e-learning", virtual and remote labs, and new technologies applied to distance education. He has published more than 200 papers both in international journals and conferences.

\end{document}